\documentclass{article}

     \PassOptionsToPackage{numbers, compress}{natbib}
\usepackage[preprint]{neurips_2024}

\usepackage[utf8]{inputenc} % allow utf-8 input
\usepackage[T1]{fontenc}    % use 8-bit T1 fonts
\usepackage{hyperref}       % hyperlinks
\usepackage{url}            % simple URL typesetting
\usepackage{booktabs}       % professional-quality tables
\usepackage{amsfonts}       % blackboard math symbols
\usepackage{nicefrac}       % compact symbols for 1/2, etc.
\usepackage{microtype}      % microtypography
\usepackage{xcolor}         % colors
\usepackage{graphicx}
\usepackage{zymacros}

\usepackage{amsmath}
\usepackage{algorithm}
\usepackage{algpseudocode}
\usepackage{subfigure}
\title{Newton Informed Neural Operator for Computing Multiple Solutions of Nonlinear Partials Differential Equations}

\author{%
    Wenrui Hao\thanks{Corresponding Authors}\\
	Department of Mathematics\\
	The Pennsylvania State University, University Park,\\State College, PA, USA\\
   \texttt{wxh64@psu.edu}  \\\And
   Xinliang Liu\\
	 Computer, Electrical and Mathematical Science and Engineering Division,\\ King Abdullah University of Science and Technology,\\ Thuwal, Saudi Arabia\\\texttt{xinliang.liu@kaust.edu.sa}
  \And
   Yahong Yang\\
	Department of Mathematics\\
	The Pennsylvania State University, University Park,\\State College, PA, USA\\\texttt{yxy5498@psu.edu}
}

\begin{document}

\maketitle

\begin{abstract}
  Solving nonlinear partial differential equations (PDEs) with multiple solutions using neural networks has found widespread applications in various fields such as physics, biology, and engineering. However, classical neural network methods for solving nonlinear PDEs, such as Physics-Informed Neural Networks (PINN), Deep Ritz methods, and DeepONet, often encounter challenges when confronted with the presence of multiple solutions inherent in the nonlinear problem. These methods may encounter ill-posedness issues.
In this paper, we propose a novel approach called the Newton Informed Neural Operator, which builds upon existing neural network techniques to tackle nonlinearities. Our method combines classical Newton methods, addressing well-posed problems, and efficiently learns multiple solutions in a single learning process while requiring fewer supervised data points compared to existing neural network methods. 
\end{abstract}

\section{Introduction}
Neural networks have been extensively applied to solve partial differential equations (PDEs) in various fields, including biology, physics, and materials science \cite{lagaris1998artificial,han2018solving}. While much of the existing work focuses on PDEs with a singular solution, nonlinear PDEs with multiple solutions pose a significant challenge but are widely encountered in applications such as \cite{amann1979multiplicity,breuer2003multiple,pearson1993complex}. In this paper, we aim to solve the following nonlinear PDEs that may contain multiple solutions:
\begin{equation}
\begin{cases}
\fL u(\vx) = f(u), & \vx\in\Omega \\
u(\vx) = 0, & \vx\in\partial\Omega
\end{cases}\label{nonlinear}
\end{equation}
Here, $\Omega$ is the domain of equation, \(f(u)\) is a nonlinear function in \(\sR\), \(u:\sR^d\to\sR\) and $\fL$ is a second-order elliptic operator given by \[\quad \fL u=-\sum_{i, j=1}^d a^{i j}(\vx) u_{x_i x_j}+\sum_{i=1}^d b^i(\vx) u_{x_i}+c(\vx) u,\] for given coefficient functions $a^{i j}, b^i, c(i, j=1, \ldots, d)$ with \(\sum_{i, j=1}^n a^{i j}(\vx) \xi_i \xi_j \geq \lambda|\vxi|^2,~\text{for a constant $\lambda\ge 0$}.\)

Various neural network methods have been developed to tackle partial differential equations (PDEs), including PINN \cite{raissi2019physics}, the Deep Ritz method \cite{yu2018deep}, DeepONet \cite{lu2019deeponet}, FNO \cite{li2020fourier}, MgNO \cite{he2023mgno}, and OL-DeepONet \cite{lin2023operator}. These methods can be broadly categorized into two types: function learning and operator learning approaches. In function learning, the goal is to directly learn the solution. However, these methods often encounter the limitation of only being able to learn one solution in each learning process. Furthermore, the problem becomes ill-posed when there are multiple solutions. On the other hand, operator learning aims to approximate the map between parameter functions in PDEs and the unique solution. This approach cannot address the issue of multiple solutions or find them in a single training session. We will discuss this in more detail in the next section.

In this paper, we present a novel neural network approach for solving nonlinear PDEs with multiple solutions. Our method is grounded in operator learning, allowing us to capture multiple solutions within a single training process, thus overcoming the limitations of function learning methods in neural networks. Moreover, we enhance our network architecture by incorporating traditional Newton methods \cite{ulbrich2002semismooth,amann1979multiplicity}, as discussed in the next section. This integration ensures that the problem of operator learning becomes well-defined, as Newton methods provide well-defined locally, thereby ensuring a robust operator. This approach addresses the challenges associated with directly applying operator networks to such problems. Additionally, we leverage Newton information during training, enabling our method to perform effectively even with limited supervised data points. We introduce our network as the \textbf{Newton Informed Neural Operator}.

As mentioned earlier, our approach combines the classical Newton method, which translates nonlinear PDEs into an iterative process involving solving linear functions at each iteration. One key advantage of our method is that, once the operator is effectively learned, there's no need to solve the linear equation in every iteration. This significantly reduces the time cost, especially in complex systems encountered in fields such as material science, biology, and chemistry.
Above all, our network, termed the Newton Informed Neural Operator, can efficiently handle nonlinear PDEs with multiple solutions. It tackles well-posed problems, learns multiple solutions in a single learning process, requires fewer supervised data points compared to existing neural network methods, and saves time by eliminating the need to repeatedly learn and solve the inverse problem as in traditional Newton methods.

The following paper is organized as follows: In the next section (Section \ref{back}), we will delve into nonlinear PDEs with multiple solutions and discuss related works on solving PDEs using neural network methods. In Section \ref{NINO}, we will review the classical Newton method for solving PDEs and introduce the Newton Informed Neural Operator, which combines neural operators with the Newton method to address nonlinear PDEs with multiple solutions. In this section, we will also analyze the approximation and generalization errors of the Newton Informed Neural Operator. Finally, in Section \ref{experiments}, we present the numerical results of our neural networks for solving nonlinear PDEs.

\section{Backgrounds and Relative Works}\label{back}

\subsection{Nonlinear PDEs with multiple solutions}

Significant mathematical models depicting natural phenomena in biology, physics, and materials science are rooted in nonlinear partial differential equations (PDEs) \cite{cross1993pattern}. These models, characterized by their inherent nonlinearity, present complex multi-solution challenges. Illustrative examples include string theory in physics, reaction-diffusion systems in chemistry, and pattern formation in biology \cite{kondo2010reaction,hao2020spatial}. However, experimental techniques like synchrotronic and laser methods can only observe a subset of these multiple solutions. Thus, there is an urgent need to develop computational methods to unravel these nonlinear models, offering deeper insights into the underlying physics and biology \cite{hoyle2006pattern}. Consequently, efficient numerical techniques for identifying these solutions are pivotal in understanding these intricate systems. Despite recent advancements in numerical methods for solving nonlinear PDEs, significant computational challenges persist for large-scale systems. Specifically, the computational costs of employing Newton and Newton-like approaches are often prohibitive for the large-scale systems encountered in real-world applications. In response to these challenges \cite{kelley2003solving}, we propose an operator learning approach based on Newton's method to efficiently compute multiple solutions of nonlinear PDEs.
 
%In this paper, we focus on the nonlinear elliptic equation  by utilized the DeepONet \cite{lu2019deeponet} to achieve our task. Our method can be generalized to other types of equations and utilizing different neural operators such as MgNO \cite{he2023mgno}, FNO \cite{li2020fourier}, etc., which we consider as future work.

\subsection{Related works}
Indeed, there are numerous approaches to solving partial differential equations (PDEs) using neural networks. Broadly speaking, these methods can be categorized into two main types: function learning and operator learning.

In function learning, neural networks are used to directly approximate the solutions to PDEs. %This involves training the neural network to approximate the solution of PDEs directly. 
Function learning approaches aim to directly learn the solution function itself. On the other hand, in operator learning, the focus is on learning the operator that maps input parameters to the solution of the PDE. Instead of directly approximating the solution function, the neural network learns the underlying operator that governs the behavior of the system.
\paragraph{Function learning methods} In the function learning, a commonly employed method for addressing this problem involves the use of Physics-Informed Neural Network (PINN)-based learning approaches, as introduced by Raissi et al. \cite{raissi2019physics}, and Deep Ritz Methods \cite{yu2018deep}. 
%However, these methods often face challenges in generating numerical solutions with high accuracy, primarily due to the potential existence of multiple solutions for the equations and the associated difficulties in achieving effective training. 
However, in these methods, the task becomes particularly challenging due to the ill-posed nature of the problem arising from multiple solutions. Despite employing various initial data and training methods, attaining high accuracy in solution learning remains a complex endeavor. Even when a high accuracy solution is achieved, each learning process typically results in the discovery of only one solution. The specific solution learned by the neural network is heavily influenced by the initial conditions and training methods employed. However, discerning the relationships between these factors and the learned solution remains a daunting task.
In \cite{huang2022hompinns}, the authors introduce a method called HomPINNs for learning multiple solution PDEs. In their approach, the number of solutions that can be learned depends on the availability of supervised data. However, for solutions with insufficient supervised data, whether HomPINN can successfully learn them or not remains uncontrollable.
\paragraph{Operator learning methods}Various approaches have been developed for operator learning to solve PDEs, including DeepONet \cite{lu2019deeponet}, which integrates physical information \cite{goswami2022physics, lin2023operator}, as well as techniques like FNO \cite{li2020fourier} inspired by spectral methods, and MgNO \cite{he2023mgno}, HANO \cite{liu2024mitigating}, and WNO \cite{li2021physicsinformed} based on multilevel methods, and transformer-based neural operators \cite{cao2021choose, guo2022transformer}. These methods focus on approximating the operator between the parameters and the solutions. Firstly, they require the solutions of PDEs to be unique; otherwise, the operator is not well-defined. Secondly, they focus on the relationship between the parameter functions and the solution, rather than the initial data and multiple solutions.

%To tackle this problem, we employ Newton's method by linearizing the equation. Given any initial guess \(u_0(\vx)\), for \(i=1,2,\ldots,M\), in the \(i\)-th iteration of Newton's method, we have
%\(\tilde{u}(\vx) = u + \delta u(\vx)\) by solving
%\begin{equation}
%\begin{cases}
%(\fL - f'(u)) \delta u(\vx) = \fL u + f(u), & \vx\in\Omega \\
%\delta u(\vx) = 0, & \vx\in\partial\Omega.
%\end{cases}
%\label{linear}
%\end{equation}
%To solve Eq.(\ref{linear}) and repeat it $M$ times will yield one of the solutions of the nonlinear equation (\ref{nonlinear}). For the differential initial data, it will converge to the differential solution of Eq.(\ref{nonlinear}), which constitutes a well-posed problem. However, applying the solver to solve Eq.~(\ref{nonlinear}) multiple times can be time-consuming, especially in high-dimensional structures or when the number of discrete points in the solver is large. In this paper, we employ neural networks to address these challenges.

\section{Newton Informed Neural Operator}\label{NINO}
\subsection{Review of Newton Methods to Solve Nonlinear Partial Differential Equations}
 To tackle this problem Eq.~\ref{nonlinear}, we employ Newton's method by linearizing the equation. Given any initial guess \(u_0(\vx)\), for \(i=1,2,\ldots,M\), in the \(i\)-th iteration of Newton's method, we have
\(\tilde{u}(\vx) = u + \delta u(\vx)\) by solving
\begin{equation}
\begin{cases}
(\fL - f'(u)) \delta u(\vx) = \fL u - f(u), & \vx\in\Omega \\
\delta u(\vx) = 0, & \vx\in\partial\Omega.
\end{cases}
\label{linear}
\end{equation}
To solve Eq.(\ref{linear}) and repeat it $M$ times will yield one of the solutions of the nonlinear equation (\ref{nonlinear}). For the differential initial data, it will converge to the differential solution of Eq.(\ref{nonlinear}), which constitutes a well-posed problem. However, applying the solver to solve Eq.~(\ref{nonlinear}) multiple times can be time-consuming, especially in high-dimensional structures or when the number of discrete points in the solver is large. In this paper, we employ neural networks to address these challenges.

%\subsection{Method}
%\begin{itemize}
%    \item \textbf{Multiple solutions}. The goal is to find a surrogate model to map the initial state $u_{int}$ to the solution. However, the existing  neural operator is black-box model, they approximate the map $\mathcal{G}:u_{int} \mapsto u_{true}$ end-to-end. In order to get all the possibility of solution,  heavily sampled data to cover all the domain of the initial conditions and solved the problem to get the true solution corresponding the sampled initial state.
%    \item  \textbf{Newton informed iterative neural operator instead of end-to-end} Instead of using end-to-end surrogate model, we use Newton informed neural operator. We propose the model to such that $\mathcal{N}: u_{t}\mapsto u_{t+1}$ and $u_{t} \rightarrow u_{true}$ as $t\rightarrow \infty$. The sequence is informed by the using iterative method to solve the problem with given initial condition $u_{int}$. Our model is a recurrent neural operator informed by newton's method. Put a diagram for illustration.
%    \item \textbf{Newton's loss} To avoid generate many sequences of  the newton iteration steps $\{u_0^{i}, ...,u_n^{i}\}_{i=1}^m$ to make sure the entire domain  is covered. We propose Newton loss such that we can sample the initial without solving the ground truth. Utilizing derivative will leads to better generalization error.
%\end{itemize}
\subsection{Neural Operator Structures} In this section, we introduce the structure of the DeepONet \cite{lu2019deeponet, lanthaler2022error} to approximate the operator locally in the Newton methods from Eq.(\ref{linear}), i.e., \(\fG(u) := u_*\), where \(u_*\) is the solution of Eq.(\ref{linear}), which depends on \(u\). If we can learn the operator \(\fG(u)\) well using the neural operator \(\fO(u;\vtheta)\), then for an initial function \(u_0\), assume the \(n\)-th iteration will approximate one solution, i.e., \((\fG + \text{Id})^{(n)}(u_0) \approx u^*\). Thus,
\[
(\fO + \text{Id})^{(n)}(u_0) \approx (\fG + \text{Id})^{(n)}(u_0) \approx u^*.
\]
For another initial data, we can evaluate our neural operator and find the solution directly.

Then we discuss how to train such an operator.
To begin, we define the following shallow neural operators with $p$ neurons for operators from $\fX$ to $\fY$ as
\begin{equation}\label{eq:snndef}
   \fO(a;\vtheta) = \sum_{i=1}^p \fA_i \sigma\left( \fW_ia + \fB_i \right) \quad \forall a \in \fX
\end{equation}
where $\fW_i \in \fL(\fX, \fY), \fB_i \in \fY$, $\fA_i \in \fL(\fY, \fY)$, and $\vtheta$ denote all the parameters in \(\{\fW_i,\fA_i,\fB_i\}_{i=1}^p\). 
Here, $\fL(\fX, \fY)$ denotes the set of all bounded (continuous) linear operators between $\fX$ and $\fY$, and $\sigma: \mathbb \mapsto \sR$ defines the nonlinear point-wise activation.

In this paper, we will use shallow DeepONet to approximate the Newton operator. To provide a more precise description, in the shallow neural network, $\fW_i$ represents an interpolation of operators. With proper and reasonable assumptions, we can present the following theorem to ensure that DeepONet can effectively approximate the Newton method operator. The proof will be provided in the appendix. Furthermore, MgNO is replaced by $\fW$ as a multigrid operator \cite{xu1996two}, and FNO is some kind of kernel operator; our analysis can be generalized to such cases.

Before the proof, we need to establish some assumptions regarding the input space $\fX\subset H^2(\Omega)$ of the operator and $f(u)$ in Eq.~(\ref{nonlinear}).

\begin{assump}\label{app assump}
    \textbf{(i):} For any $u\in\fX$, we have that \[f'(u)\cap\{\text{eigenvalues of }\fL\}=\emptyset.\]

    \textbf{(ii):} There exists a constant $F$ such that \(\|f(x)\|_{W^{2,\infty}(\sR)}\le F\).

\textbf{(iii):} All coefficients in $\fL$ are $C^1$ and $\partial \Omega\in C^2$.

\textbf{(iv):} There exists a linear bounded operator $\fP$ on $\fX$, a small constant $\epsilon>0$, and a constant $n$ such that\[\|u-\fP u\|_{H^2(\Omega)}\le \epsilon,~\text{for all }u\in\fX.\] Furthermore, $\fP u$ is an $n$-dimensional term, i.e., it can be denoted by the $n$-dimensional vector $\bar{\fP} u\in\sR^n$.

%\textbf{(v)}: There exists a constant $\epsilon_2$. For any linear transformation $\vW\in\sR^{n\times n}$ of $\bar{\fP}$, there exists a multigrid operator $\fW_{Mg}$ that depends on $\vW$, which can approximate $\vM\cdot \bar{\fP}$ well, i.e., \[|\fW_{Mg} u-\vM\cdot \bar{\fP} u|\le \epsilon_2,~\text{for all }u\in\fX.\] Denote $\epsilon=\max\{\epsilon_1,\epsilon_2\}$.
\end{assump}

\begin{rmk}\label{rm app}
    We want to emphasize the reasonableness of our assumptions. For condition (i), we are essentially restricting our approximation efforts to local regions. This limitation is necessary because attempting to approximate the neural operator across the entire domain could lead to issues, particularly in cases where multiple solutions exist. Consider a scenario where the input function $u$ lies between two distinct solutions. Even a small perturbation of $u$ could result in the system converging to a completely different solution. Condition (i) ensures that Equation (\ref{linear}) has a unique solution, allowing us to focus our approximation efforts within localized domains.

    Conditions (ii) and (iii) serve to regularize the problem and ensure its tractability. These conditions are indeed straightforward to fulfill, contributing to the feasibility of the overall approach.

   For the embedding operator $\fP$ in (iv), there are a lot of choices in DeepONet, such as finite element methods like Argyris elements \cite{brenner2008mathematical} or embedding methods in \cite{li2002new,hu2015minimal}. We will discuss more in the appendix. The approximation order would be $h^{-(l-2)}$, and the differential method may achieve a different order. We omit the detailed discussion in the paper. Furthermore, for the differential neural network, this embedding may be different; for example, we can use Fourier expansion \cite{yang2022approximation} or multigrid methods \cite{he2023mgno} to achieve this task.

    %The flexibility in choosing $\fP$ in condition (iv) allows for tailoring the method to the specific problem at hand, whether it be through interpolation, Fourier expansion, or other differential methods. This adaptability accommodates varying degrees of accuracy and computational efficiency depending on the chosen approach.

    %Condition (v) leverages the property of multigrid operators to approximate elliptic operators effectively, a characteristic essential to the success of the method. The accuracy of the approximation may vary depending on the specific equation, input space $\fX$, and discrete method $\fP$. Exploring these nuances and optimizing the method accordingly presents an avenue for future research and refinement.
\end{rmk}

\begin{thm}\label{approximation}
    Suppose $\fX=\fY\subset H^2(\Omega)$ and Assumption \ref{app assump} holds. Then, there exists a neural network $\fO(u;\vtheta)\in \Xi_p$ defined as
 \begin{equation}
    \Xi_p:=\left\{\sum_{i=1}^{p} \vA_{i} \sigma\left(\fW_iu+\vb_i\right) \sigma\left( \vw_{i} \cdot \vx + \zeta_i \right)|  \fW_i \in \fL(\fX, \sR^m), \vb_i \in \sR^m, \vA_i \in \fL(\sR^m, \sR)   \right\}
\end{equation} 
such that
\begin{equation}
\left\|\sum_{i=1}^{p} \vA_{i} \sigma\left(\fW_iu+\vb_i\right) \sigma\left( \vw_{i} \cdot \vx + \zeta_i \right)-\fG(u)\right\|_{L^2(\Omega)}\le C_1m^{-\frac{1}{ n}}+C_2(\epsilon+p^{-\frac{2}{d}}),
\end{equation}
where $\sigma$ is a smooth non-polynomial activation function, $C_1$ is a constant independent of $m$, $\epsilon$, and $p$, $C_2$ is a constant depended on $p$, $n$ is the scale of the $\fP$ in Assumption \ref{app assump}. And $\epsilon$ depends on $n$. $n$ and $\epsilon$ are defined in Assumption \ref{app assump}.
\end{thm}

\subsection{Loss Functions of Newton Informed Neural Operator}
\subsubsection{Mean Square Error}
The Mean Square Error loss function is defined as:
\begin{align}
\fE_{S}(\vtheta):=\frac{1}{M_u\cdot M_x} \sum_{j=1}^{M_u} \sum_{k=1}^{M_x} \left|\fG\left(u_j\right)\left(\vx_k\right)-\fO\left(u_j;\vtheta\right)\left(\vx_k\right)\right|^2
\label{eqn:mse_loss}
\end{align}
where $u_1,u_2,\ldots,u_{M_u}\sim\mu$ are independently and identically distributed (i.i.d) samples in $\mathcal{X}$, and $\vx_1,\vx_2,\ldots,\vx_{M_x}$ are uniformly i.i.d samples in $\Omega$. %Denote \[\vtheta_1:=\arg\min \fE_S(\vtheta).\]

However, using only the Mean Squared Error loss function is not sufficient for training to learn the Newton method, especially since in most cases, we do not have enough data $\{u_i,\fG\left(u_j\right)\}_{j=1}^{M_u}$. Furthermore, there are situations where we do not know how many solutions exist for the nonlinear equation (\ref{nonlinear}). If the data is sparse around one of the solutions, it becomes impossible to effectively learn the Newton method around that solution.

Given that $\fE_{S}(\vtheta)$ can be viewed as the finite data formula of $\fE_{Sc}(\vtheta)$, where \[\fE_{Sc}(\vtheta)=\lim_{M_u,M_x\to\infty}\fE_{S}(\vtheta).\] The smallness of $\fE_{Sc}$ can be inferred from Theorem \ref{approximation}. To understand the gap between $\fE_{Sc}(\vtheta)$ and $\fE_{S}(\vtheta)$, we can rely on the following theorem. Before the proof, we need some assumptions about the data in $\fE_{S}(\vtheta)$:
\begin{assump} \label{gen app}
\textbf{(i) Boundedness:} For any neural network with bounded parameters, characterized by a bound $B$ and dimension $d_{\vtheta}$, there exists a function $\Psi: L^2(\Omega) \rightarrow[0, \infty)$ such that
\[
|\fG(u)(\vx)| \leqslant \Psi(u), \quad \sup _{\vtheta \in[-B, B]^{d_{\vtheta}}}\left|\fN(u;\vtheta)(\vx)\right| \leqslant \Psi(u),
\]
for all $u \in \fX, \vx \in \Omega$, and there exist constants $C, \kappa>0$, such that
\begin{align}
\Psi(u) \leqslant C\left(1+\|u\|_{H^2}\right)^\kappa .\label{kappa}
\end{align}

\textbf{(ii) Lipschitz continuity:} There exists a function $\Phi: L^2(\Omega) \rightarrow[0, \infty)$, such that
\begin{align}
\left|\fN(u;\vtheta)(\vx)-\fN(u;\vtheta')(\vx)\right| \leqslant \Phi(u)\left\|\vtheta-\vtheta^{\prime}\right\|_{\ell^{\infty}}
\end{align}
for all $u \in \fX, \vx \in \Omega$, and
\[
\Phi(u) \leqslant C\left(1+ \|u \|_{H^2(\Omega)}\right)^\kappa,
\]
for the same constants $C, \kappa>0$ as in Eq.~(\ref{kappa}).

 \textbf{(iii) Finite measure:} There exists $\alpha>0$, such that
\[
\int_{H^2(\Omega)} e^{\alpha\|u\|_{H^2(\Omega)}^2} \D \mu(u)<\infty .
\]
 \end{assump}

\begin{thm}\label{genL2} 
If Assumption \ref{gen app} holds, then the generalization error is bounded by
\[
|\rmE(\fE_{S}(\vtheta) - \fE_{Sc}(\vtheta))| \leqslant C\left[\frac{1}{\sqrt{M_u}}\left(1 + C d_{\vtheta} \log (C B \sqrt{M_u})^{2 \kappa + 1 / 2}\right) + \frac{d_{\vtheta}\log M_x}{M_x}\right],
\]
where $C$ is a constant independent of $B$, $d_{\vtheta}$, $M_x$, and $M_u$. The parameter $\kappa$ is specified in (\ref{kappa}). Here, $B$ represents the bound of parameters and $d_{\vtheta}$ is the number of parameters.
\end{thm}
The proof of Theorem \ref{genL2} is presented in Appendix \ref{gen}.

\begin{rmk}
    The Assumption \ref{gen app} is easy to achieve if we consider $\fX$ as the local function set around the solution, which is typically the case in Newton methods. This aligns with our approach and working region in the approximation part (see Remark \ref{rm app}).
\end{rmk}

%$\fG[u(\vx)] = \delta u(\vx)$. We focus solely on learning $\delta u(\vx)$. As $\fG[u(\vx)]$ represents the exact result from Newton's method, this MSE loss functions as supervised training loss. It's important to note that this method requires computing all $\fG\left(u_j\right)\left(\vy_k^j\right)$ to calculate the loss.

\subsubsection{Newton Loss}
As we have mentioned, relying solely on the MSE loss function can require a significant amount of data to achieve the task. However, obtaining enough data can be challenging, especially when the equation is complex and the dimension of the input space is large. Hence, we need to consider another loss function to aid learning, which is the physical information loss function \cite{raissi2019physics,goswami2022physics,huang2022hompinns,li2021physicsinformed}, referred to here as the Network loss function. 

The Newton loss function is defined as:
\begin{align}
\fE_{N}(\vtheta):=\frac{1}{N_u\cdot N_x} \sum_{j=1}^{N_u} \sum_{k=1}^{N_x} \left|(\fL - f'(u_j\left(\vx_k\right))) \fO\left(u_j;\vtheta\right)\left(\vx_k\right) - \fL u_j\left(\vx_k\right) - f(u_j\left(\vx_k\right))\right|^2\label{eqn:newton loss}
\end{align}
where $u_1,u_2,\ldots,u_{N_u}\sim\nu$ are independently and identically distributed (i.i.d) samples in $\mathcal{X}$, and $\vx_1,\vx_2,\ldots,\vx_{N_x}$ are uniformly i.i.d samples in $\Omega$.

Given that $\fE_{N}(\vtheta)$ can be viewed as the finite data formula of $\fE_{Nc}(\vtheta)$, where \[\fE_{Nc}(\vtheta)=\lim_{N_u,N_x\to\infty}\fE_{S}(\vtheta).\] To understand the gap between $\fE_{Nc}(\vtheta)$ and $\fE_{N}(\vtheta)$, we can rely on the following theorem:

\begin{cor}\label{genH2} 
If Assumption \ref{gen app} holds, then the generalization error is bounded by
\[
|\rmE(\fE_{N}(\vtheta) - \fE_{Nc}(\vtheta))| \leqslant C\left(\frac{d_{\vtheta}\log N_u}{\sqrt{N_u}} + \frac{d_{\vtheta}\log N_x}{N_x}\right),
\]
where $C$ is a constant independent of $B$, $d_{\vtheta}$, $M_x$, and $M_u$. The parameter $\kappa$ is specified in (\ref{kappa}). Here, $B$ represents the bound of parameters and $d_{\vtheta}$ is the number of parameters.
\end{cor}

The proof of Corollary \ref{genH2} is similar to that of Theorem \ref{genL2}; therefore, it will be omitted from the paper.

\begin{rmk}
    If we only utilize $\fE_{S}(\vtheta)$ as our loss function, as demonstrated in Theorem \ref{genL2}, we require both $M_u$ and $M_x$ to be large, posing a significant challenge when dealing with complex nonlinear equations. Obtaining sufficient data becomes a critical issue in such cases. In this paper, we integrate Newton information into the loss function, defining it as follows:
\begin{equation}
    \fE(\vtheta):=\lambda\fE_{S}(\vtheta)+\fE_{N}(\vtheta),
    \label{eqn:regularizedloss}
\end{equation}
where $\fE_{N}(\vtheta)$ represents the cost of the data involved in unsupervised learning. If we lack sufficient data for $\fE_{S}(\vtheta)$, we can adjust the parameters by selecting a small $\lambda$ and increasing $N_x$ and $N_u$. This strategy enables effective learning even when data for $\fE_{S}(\vtheta)$ is limited.

\end{rmk}

In the following experiment, we will use the neural operator established in Eq.~(\ref{eq:snndef}) and the loss function in Eq.~(\ref{eqn:regularizedloss}) to learn one step of the Newton method locally, i.e., the map between the input \(u\) and the solution \(\delta u\) in eq.~(\ref{linear}). If we have a large dataset, we can choose a large \(\lambda\) in \(\fE(\vtheta)\) (\ref{eqn:regularizedloss}); if we have a small dataset, we will use a small \(\lambda\) to ensure the generalization of the operator is minimized. After learning one step of the Newton method using the operator neural networks, we can easily and quickly obtain the solution by the initial condition of the nonlinear PDEs (\ref{nonlinear}) and find new solutions not present in the datasets.
%\paragraph{Discrete Newton Loss}

\section{Experiments}\label{experiments}
\subsection{Experimental Settings}

We introduce two distinct training methodologies. The first approach employs exclusively supervised data, leveraging the Mean Squared Error Loss \eqref{eqn:mse_loss} as the primary optimization criterion. The second method combines both supervised and unsupervised learning paradigms, utilizing a hybrid  loss function \ref{eqn:regularizedloss} that integrates Mean Squared Error Loss \eqref{eqn:mse_loss} for small proportion of data with ground truth (supervised training dataset) and with Newton's loss \eqref{eqn:newton loss} for large proportion of data without ground truth (unsupervised training dataset). We call the two methods as \textbf{method 1} and \textbf{method 2}. The approaches are designed to simulate a practical scenario with limited data availability, facilitating a comparison between these training strategies to evaluate their efficacy in small supervised data regimes. We choose the same configuration of neural operator (DeepONet) which is align with our theoretical analysis. One can find the detailed experimental settings and the datasets for each examples below in Appendix \ref{sec:experiment_settings}.

\subsection{Case 1: Convex problem}
We consider 2D convex problem $\mathcal{L}(u)-f(u)=0$, where  $\mathcal{L}(u):= -\Delta u$,  $f(u):-u^2 + \sin 5\pi (x+y)$ and $u=0$ on $\partial \Omega$. We investigate the training dynamics and testing performance of neural operator (DeepONet) trained with different loss functions and dataset sizes, focusing on Mean Squared Error (MSE) and Newton's loss functions.
 
\begin{figure}[ht]
\centering
\subfigure[The training and testing errors using method 1]{
    \includegraphics[width=0.4\textwidth]{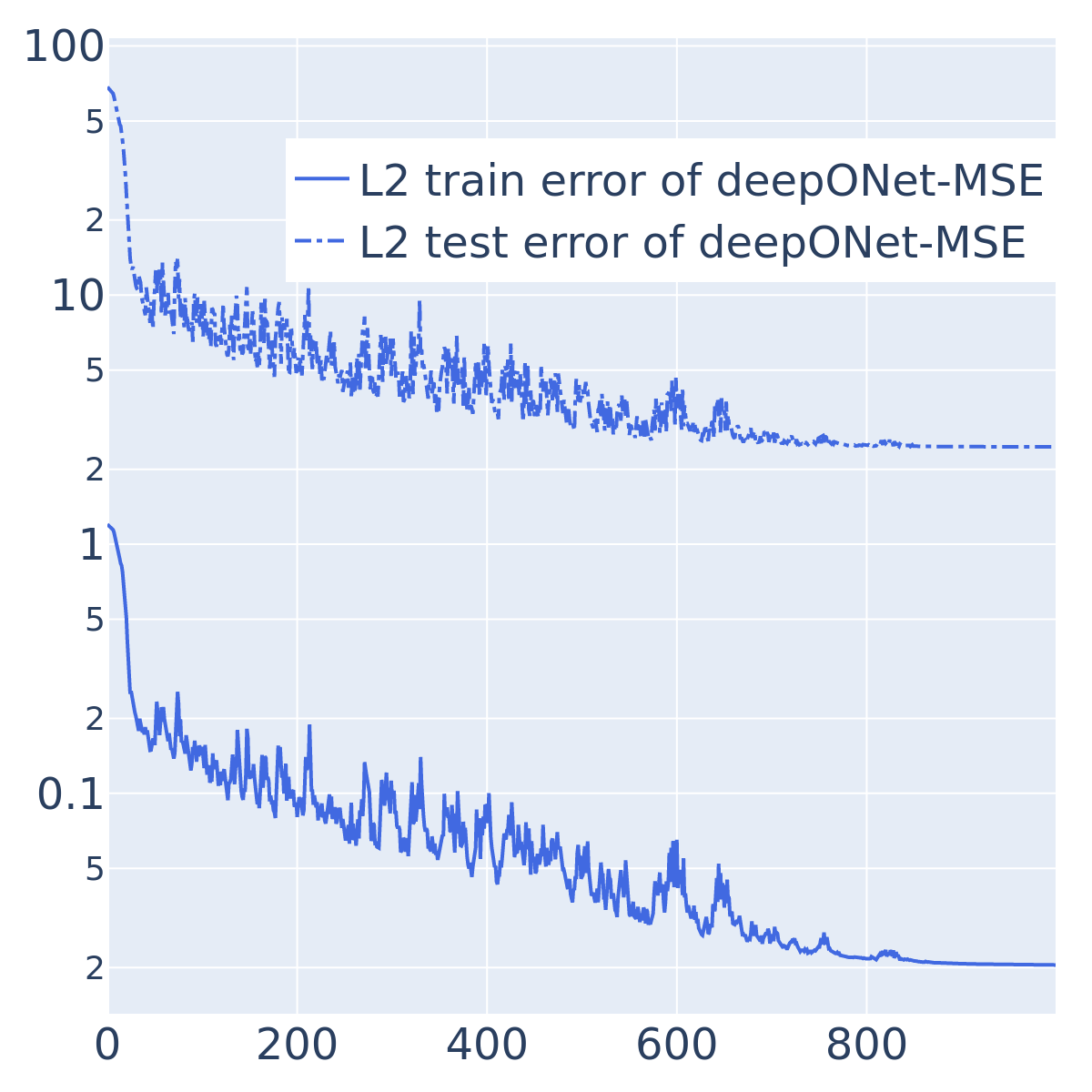}
    \label{fig:mse_small}
}
% \subfigure[Newton's loss on a large dataset(5000) without ground truth(label)]{
%     \includegraphics[width=0.3\textwidth]{figure/deepONet-Newton-l2.png}
%     \label{fig:newton_large}
% }
\subfigure[Comparison of models trained using Method 1 and Method 2, the latter employing the regularized loss function as defined in Equation \ref{eqn:regularizedloss} with $\lambda = 0.01$.
]{
    \includegraphics[width=0.4\textwidth]{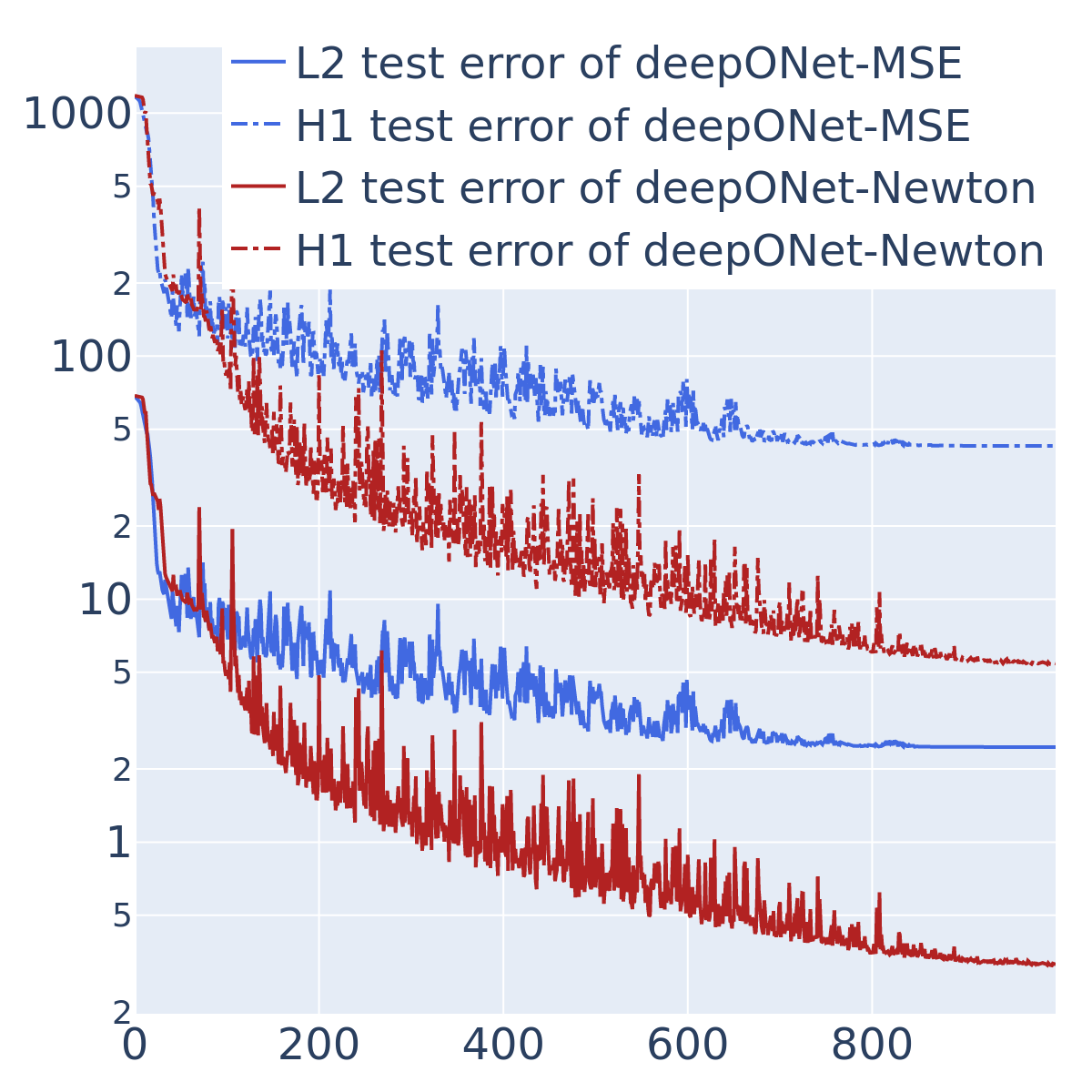}
    \label{fig:test_compare}
}
\caption{Training and testing performance of DeepONet under various conditions. For the detailed experiments settings, please refer to the Appendix \ref{sec:experiment_settings}.}
\label{fig:deepOnet}
\end{figure}

\textbf{MSE Loss Training (Fig. \ref{fig:mse_small})}: In method 1, Effective training is observed but exhibits poor generalization, suggesting only MSE loss is not enough.
%\textbf{Newton's Loss Training (Fig. \ref{fig:newton_large})}: Demonstrates improved generalization, with a steady decrease in both training and testing errors, indicative of robust learning on an extensive dataset.
\textbf{Performance Comparison (Fig. \ref{fig:test_compare})}: Newton's loss model (\textbf{method 2}) exhibits superior performance in both $L_2$ and $H_1$ error metrics, highlighting its enhanced generalization accuracy. This study shows the advantages of using Newton's loss for training DeepONet models, requiring fewer supervised data points compared to \textbf{method 1}.

\subsection{Case 2: Non-convex problem with multiple solutions}
\begin{figure}
    \centering
    \subfigure[Solutions of 2D Non-convex problem \eqref{2dex}]{
    \includegraphics[width=.6\textwidth]{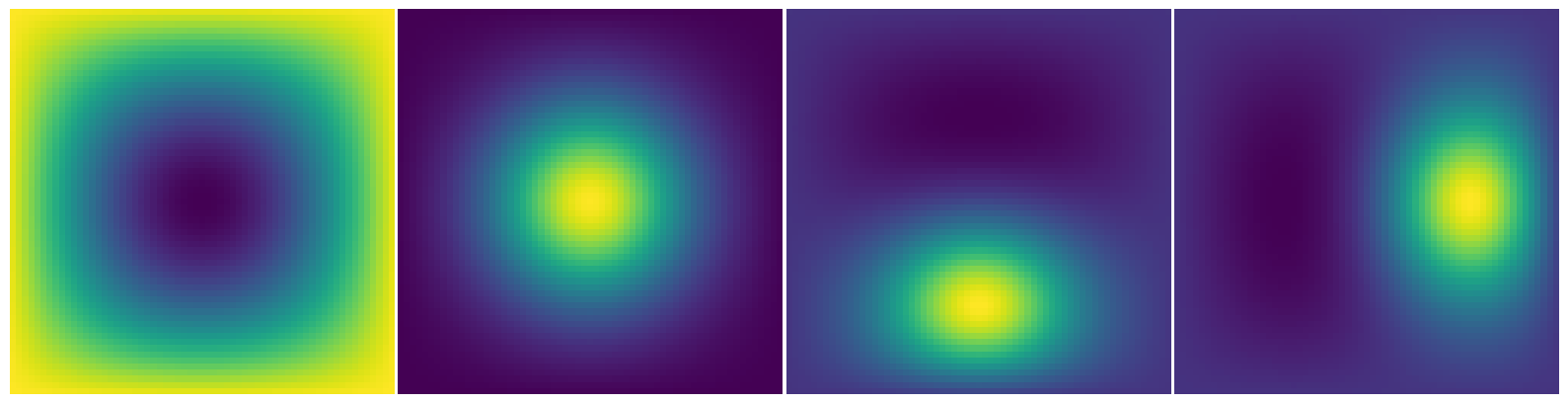}}
    \subfigure[Method 1 VS Method 2]{\includegraphics[width=.3\textwidth, height=.25\textwidth]{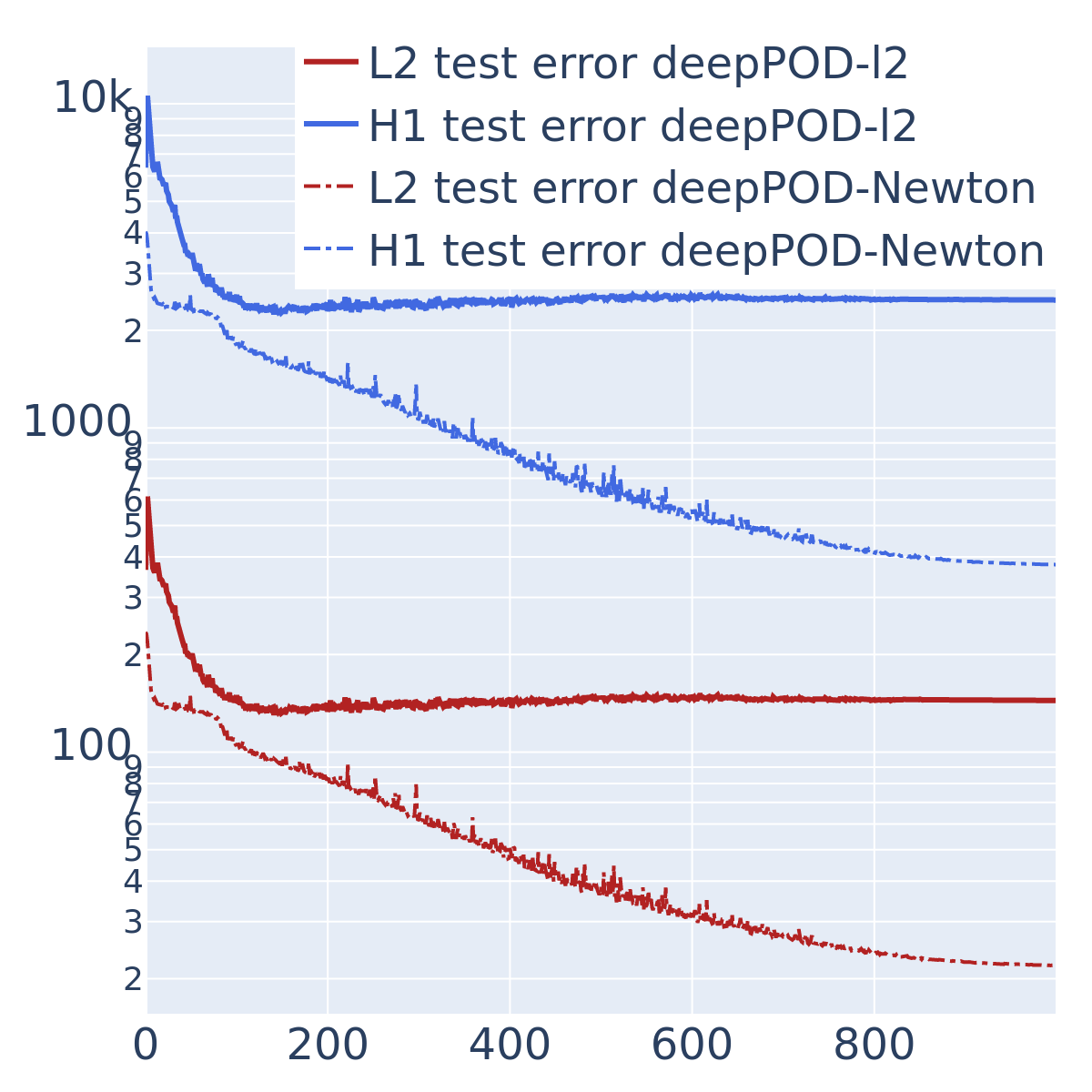}}
    \caption{Solutions of 2D Non-convex problem \eqref{2dex}}
    \label{fig:case2}
\end{figure}

We consider a 2D Non-convex problem,
\begin{equation}\label{2dex}
\begin{cases} -\Delta u(x,y)-u^2(x,y)=-s\sin(\pi x)\sin(\pi y)  \quad \text{in} \quad \Omega, \\u(x,y)=0,\quad  \text{in} \quad \partial \Omega\end{cases} 
\end{equation}
where $\Omega=(0,1) \times (0,1)$ \cite{breuer2003multiple}. 
In this case, $\mathcal{L}(u):= -\Delta(u)-u^2$ and it is non-convex, with multiple solutions (see Figure\ref{fig:case2} for its solutions).

In the experiment, we let one of the multiple ground truth solutions is rarely touched in the supervised training dataset such that the neural operator trained via \textbf{method 1} will saturate in terms of test error because of the it relies on the ground truth to recover all the patterns for multiple solution cases (as shown by the curves in Figure \ref{fig:case2}). On the other hand, the model trained via \textbf{method 2} is less affected by the limited supervised data since the utilization of the Newton's loss.  One can refer Appendix \ref{sec:experiment_settings} for the detailed experiments setting. 

\paragraph{Efficiency}
We utilize this problem as a case study to demonstrate the superior efficiency of our neural operator-based method as a surrogate model to Newton' method. By benchmarking both approaches, we highlight the significant performance advantages of our neural operator model. The performance was assessed in terms of total execution time, which includes the setup of matrices and vectors, computation on the GPU, and synchronization of CUDA streams. Both methods leverage the parallel processing capabilities of the GPU. Specifically, the Newton solver explicitly uses 10 streams and CuPy with CUDA to parallelize the computation and fully utilize the GPU parallel processing capabilities, aiming to optimize execution efficiency. In contrast, the neural operator method is inherently parallelized, taking full advantage of the GPU architecture without the explicit use of multiple streams as indicated in the table. The computational times of both methods were evaluated under a common hardware configuration. One can find the detailed description of the experiments in \ref{sec:bench}.

\begin{table}[ht]
\centering
\begin{tabular}{@{}lcc@{}}
\toprule
\textbf{Parameter} & \textbf{Newton's Method} & \textbf{Neural Operator } \\ \midrule
Number of Streams & 10 & - \\
Data Type & float32 & float32 \\
Execution Time for 500 linear Newton systems (s) & 31.52 & 1.1E-4 \\
Execution Time for 5000 linear Newton systems (s) & 321.15 & 1.4E-4 \\
\bottomrule
\end{tabular}
\caption{Benchmarking the efficiency of the neural operator-based method}
\label{tab:benchmark}
\end{table}

The data presented in the table illustrates that the neural operator is significantly more efficient than the classical Newton solver on GPU. This efficiency gain is likely due to the neural operator's ability to leverage parallel processing capabilities of GPUs more effectively than the Newton solver, though the Newton solver is also parallelized. This enhancement is crucial to improves efficiency for calculating the vast amounts of unknown patterns for complex nonlinear system such as Gray Scott model.

\subsection{Gray-Scott Model}
The Gray-Scott model \cite{pearson1993complex} describes the reaction and diffusion of two chemical species, \(A\) and \(S\), governed by the following equations:
\begin{align*}
\frac{\partial A}{\partial t} &= D_A \Delta A - S A^2 + (\mu + \rho) A, \\
\frac{\partial S}{\partial t} &= D_S \Delta S + S A^2 - \rho (1 - S),
\end{align*}
where \(D_A\) and \(D_S\) are the diffusion coefficients, and \(\mu\) and \(\rho\) are rate constants. 
\paragraph{Newton's Method for Steady-State Solutions}
Newton's method is employed to find steady-state solutions (\(\frac{\partial A}{\partial t} = 0\) and \(\frac{\partial S}{\partial t} = 0\)) by solving the nonlinear system:
\begin{equation}
    \begin{aligned}
0 &= D_A \Delta A - S A^2 + (\mu + \rho) A, \\
0 &= D_S \Delta S + S A^2 - \rho (1 - S).
\end{aligned}
\end{equation}

The Gray-Scott model is highly sensitive to initial conditions, where even minor perturbations can lead to vastly different emergent patterns. Please refer Figure \ref{fig:As} for some examples of the patterns. This sensitivity reflects the model's complex, non-linear dynamics that can evolve into a multitude of possible steady states based on the initial setup. Consequently, training a neural operator to map initial conditions directly to their respective steady states presents significant challenges. Such a model must  learn from a vast functional space, capturing the underlying dynamics that dictate the transition from any given initial state to its final pattern. This complexity and diversity of potential outcomes is the inherent difficulty in training neural operators effectively for systems as complex as the Gray-Scott model. One can refer \ref{sec:Gray} for the detailed discussion on Gray-Scott model. We employ a Neural Operator as a substitute for the Newton solver in the Gray-Scott model, which recurrently maps the initial state to the steady state.
\vspace{-4pt}
\begin{figure}[ht]
\centering
\subfigure[An example demonstrating how the neural operator maps the initial state to the steady state in a step-by-step manner]{\includegraphics[width=.8\textwidth]{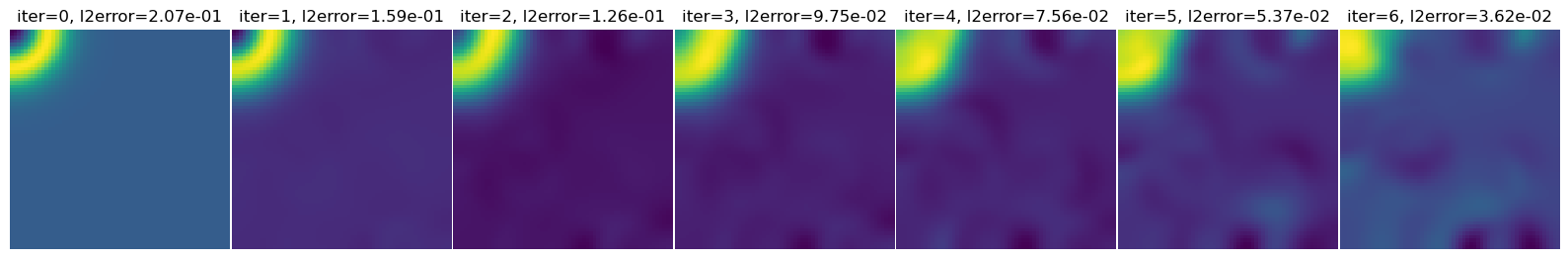}}
\vspace{-3pt}
\subfigure[Average convergence rate of.]{\includegraphics[width=.4\textwidth]{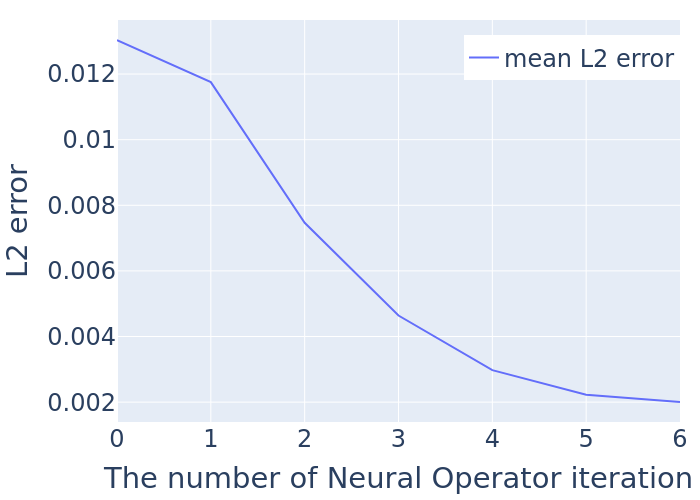}}
\subfigure[Training via method 2]{\includegraphics[width=.4\textwidth, height=.3\textwidth]{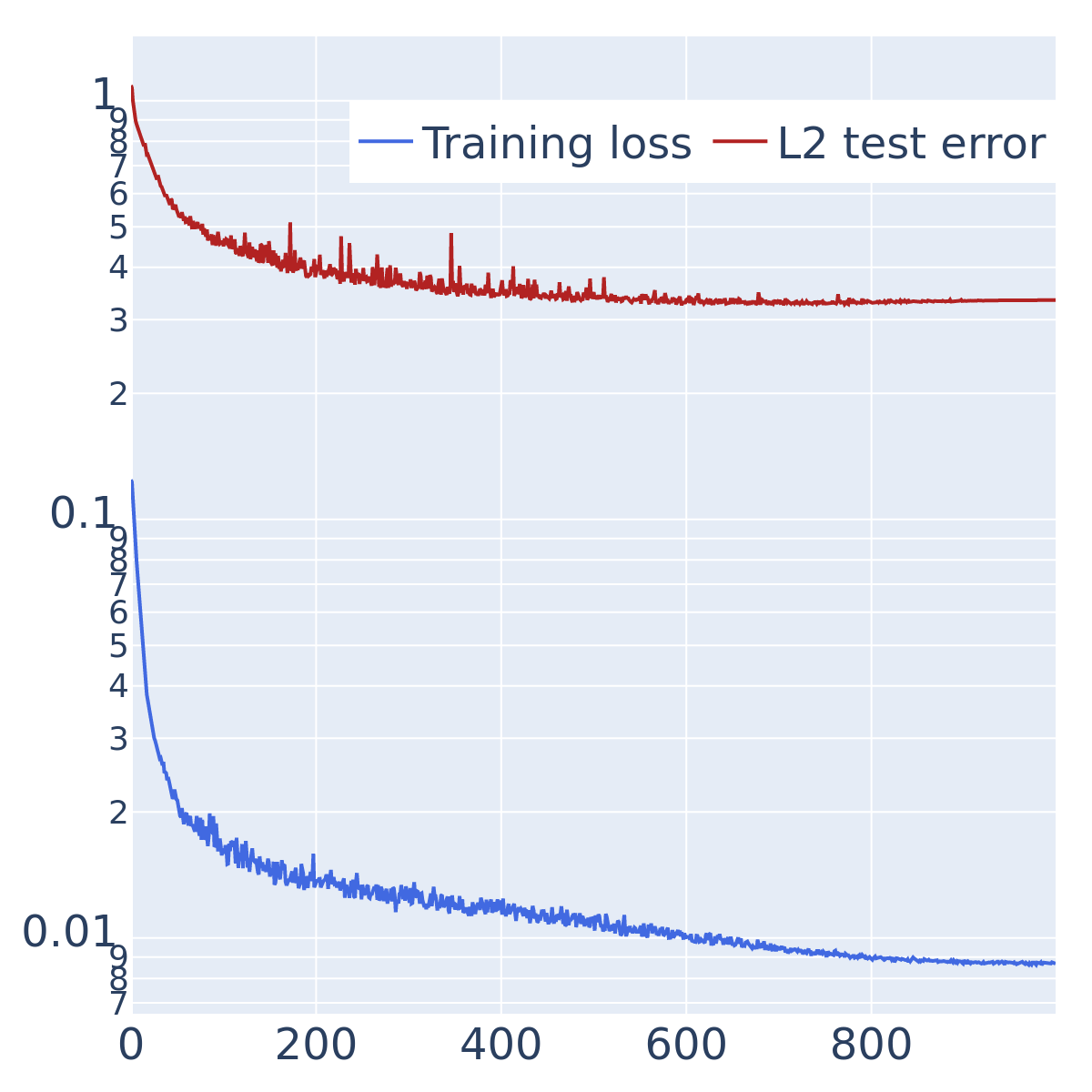}}
\caption{The convergence behavior of the Neural Operator-based solver.}
\label{fig:As}
\end{figure}

In subfigure (a), we use a ring-like pattern as the initial state to test our learned neural operator. This particular pattern does not appear in the supervised training dataset and lacks corresponding ground truth data. Despite this, our neural operator, trained using Newton's loss, is able to approximate the mapping of the initial solution to its correct steady state effectively. we further test our neural operator, utilizing it as a surrogate for Newton's method to address nonlinear problems with an initial state drawn from the test dataset. The curve shows the average convergence rate of $\|u-u_i\|$ across the test dataset, where $u_i$ represents the prediction at the $i$-th step by the neural operator. In subfigure (c), both the training loss curves and the test error curves are displayed, illustrating their progression throughout the training period.

\section{Conclusion}
In this paper, we consider using neural operators to solve nonlinear PDEs (Eq.~(\ref{nonlinear})) with multiple solutions. To make the problem well-posed and to learn the multiple solutions in one learning process, we combine neural operator learning with classical Newton methods, resulting in the Newton informed neural operator. We provide a theoretical analysis of our neural operator, demonstrating that it can effectively learn the Newton operator, reduce the number of required supervised data, and learn solutions not present in the supervised learning data due to the addition of the Newton loss (\ref{eqn:newton loss}) in the loss function. Our experiments are consistent with our theoretical analysis, showcasing the advantages of our network as mentioned earlier, i.e., it requires fewer supervised data, learns solutions not present in the supervised learning data, and costs less time than classical Newton methods to solve the problem.

\paragraph{Reproducibility Statement} 
Code Availability: The code used in our experiments can be accessed via \url{https://github.com/xll2024/newton/tree/main} and also the supplementary material. Datasets can be downloaded via urls in the repository. This encompasses all scripts, functions, and auxiliary files necessary to reproduce our results. Configuration Transparency: All configurations, including hyperparameters, model architectures, and optimization settings, are explicitly provided in Appendix. 

% A different perspective from the residual will result in a different loss. Similar to the previous approach, our aim is to have $-\{J_F(u_i)\}^{-1}F(u_i) \approx G(u_i)$. We can modify the previous loss without computing $v_i$ as:
% \begin{align}
%    L_{\theta}(\theta) = \frac{1}{mM}\sum_{i=1}^M\sum_{j=1}^m| J_F(u_i)\{ G(u_i)(x_j;\boldsymbol{\theta})\}+F(u_i)|^2
% \end{align}
% which simplifies to:
% \begin{align}
% \label{eqn:newton loss}
%    L_{\theta}(\theta) = \frac{1}{mM}\sum_{i=1}^M\sum_{j=1}^m| J_F(u_i)G+F(u_i)|^2
% \end{align}

% This represents unsupervised training loss as $v_i$ is not computed, but the focus is on computing the residual.

% For training with the MSE loss function, we will choose the number of functions $M$ as $100$ , leading to the dataset structure $\{\{u_i(\mathbf{x}_j)\}_{j=1}^{100},\{v_i(\mathbf{x}_j)\}_{j=1}^{100}\}_{i=1}^{100}$. In contrast, for Newton's Loss, we will utilize a larger dataset with $M$ set as $1000$, resulting in the dataset structure $\{\{\{u_i(\mathbf{x}_j)\}_{j=1}^m\}_{i=1}^{1000}\}$. This disparity in data quantity arises from the computation of $v_i$ for the MSE loss, which demands smaller data, while Newton's loss, not requiring $v_i$ computation, allows for the use of a larger dataset.

\begin{ack}
Y.Y. and W.H. was supported by National Institute of General Medical Sciences through grant 1R35GM146894. The work of X.L. was partially supported by the KAUST Baseline Research Fund.
\end{ack}

\newpage
\bibliographystyle{plain}
\bibliography{references,ref}

%%%%%%%%%%%%%%%%%%%%%%%%%%%%%%%%%%%%%%%%%%%%%%%%%%%%%%%%%%%%

\appendix

\section{Experimental settings}
\label{sec:experiment_settings}
\subsection{Background on the PDEs and generation of datasets}
\subsubsection{Case 1: convex problem}

\paragraph{Function and Jacobian}
The function \( F(u) \) might typically be defined as:
\[
F(u) = -\Delta u + u^2
\]
The Jacobian \( J(u) \), for the given function \( F(u) \), involves the derivative of \( F \) with respect to \( u \), which includes the Laplace operator and the derivative of the nonlinear term:
\[
J(u) = -\Delta + 2 \cdot u.
\]

The dataset are generated by sampling the initial state $u_0 \sim \mathcal{ N}(0, \Delta^{-3})$ and then calculate the convergent sequence $\{u_0,u_1,..., u_n\}$ by Newton's method. Each convergent sequence $\{u_0,u_1,..., u_n\}$ is one data entry in the dataset.

The analysis of function and Jacobian for the non-convex problem (case 2) is similar to the convex problem except that its Jacobian $J(u)=\Delta - 2 u$ such that Newton's system is not positive definite.

\subsubsection{Gray Scott model}
\label{sec:Gray}
\paragraph{Jacobian Matrix}
The Jacobian matrix \(J\) of the system is crucial for applying Newton's method:
\[
J = \begin{bmatrix}
J_{AA} & J_{AS} \\
J_{SA} & J_{SS}
\end{bmatrix}
\]
with components:
\begin{align*}
J_{AA} &= -D_A \Delta + \text{diag}(-2 S A + \mu + \rho), \\
J_{AS} &= \text{diag}(-A^2), \\
J_{SA} &= \text{diag}(2 S A), \\
J_{SS} &= -D_S \Delta + \text{diag}(A^2 + \rho).
\end{align*}

The numerical simulation of the Gray-Scott model was configured with the following parameters:

\begin{itemize}
  \item \textbf{Grid Size}: The simulation grid is square with \( N = 63 \) points on each side, leading to a total of \( N^2 \) grid points. This resolution was chosen to balance computational efficiency with spatial resolution sufficient to capture detailed patterns.  The spacing between each grid point, \( h \), is computed as \( h = \frac{1.0}{N - 1} \). This ensures that the domain is normalized to a unit square, which simplifies the analysis and scaling of diffusion rates.
  \item \textbf{Diffusion Coefficients}: The diffusion coefficients for species \( A \) and \( S \) are set to \( D_A = 2.5 \times 10^{-4} \) and \( D_S = 5.0 \times 10^{-4} \), respectively. These values determine the rate at which each species diffuses through the spatial domain.
  \item \textbf{Reaction Rates}: The reaction rate \( \mu \) and feed rate \( \rho \) are crucial parameters that govern the dynamics of the system. For this simulation, \( \mu \) is set to 0.065 and \( \rho \) to 0.04, influencing the production and removal rates of the chemical species.
\end{itemize}

\paragraph{Simulations
}
The simulation utilizes a finite difference method for spatial discretization and Newton's method to solve the steady state given the initial state. The algorithm is detailed in \ref{alg:newton}.

\begin{figure}
    \centering
    \includegraphics[width=1\textwidth]{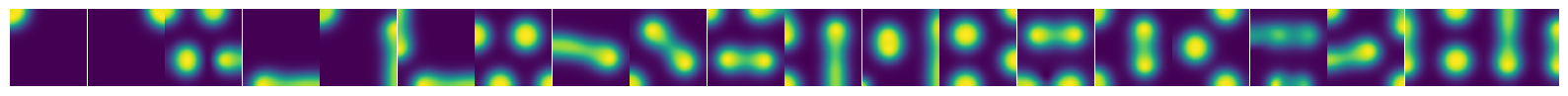}
    \caption{Examples of steady states of the Gray Scott model}
    \label{fig:As}
\end{figure}

\begin{figure}[ht]
\centering
\subfigure{
    \includegraphics[width=0.3\textwidth]{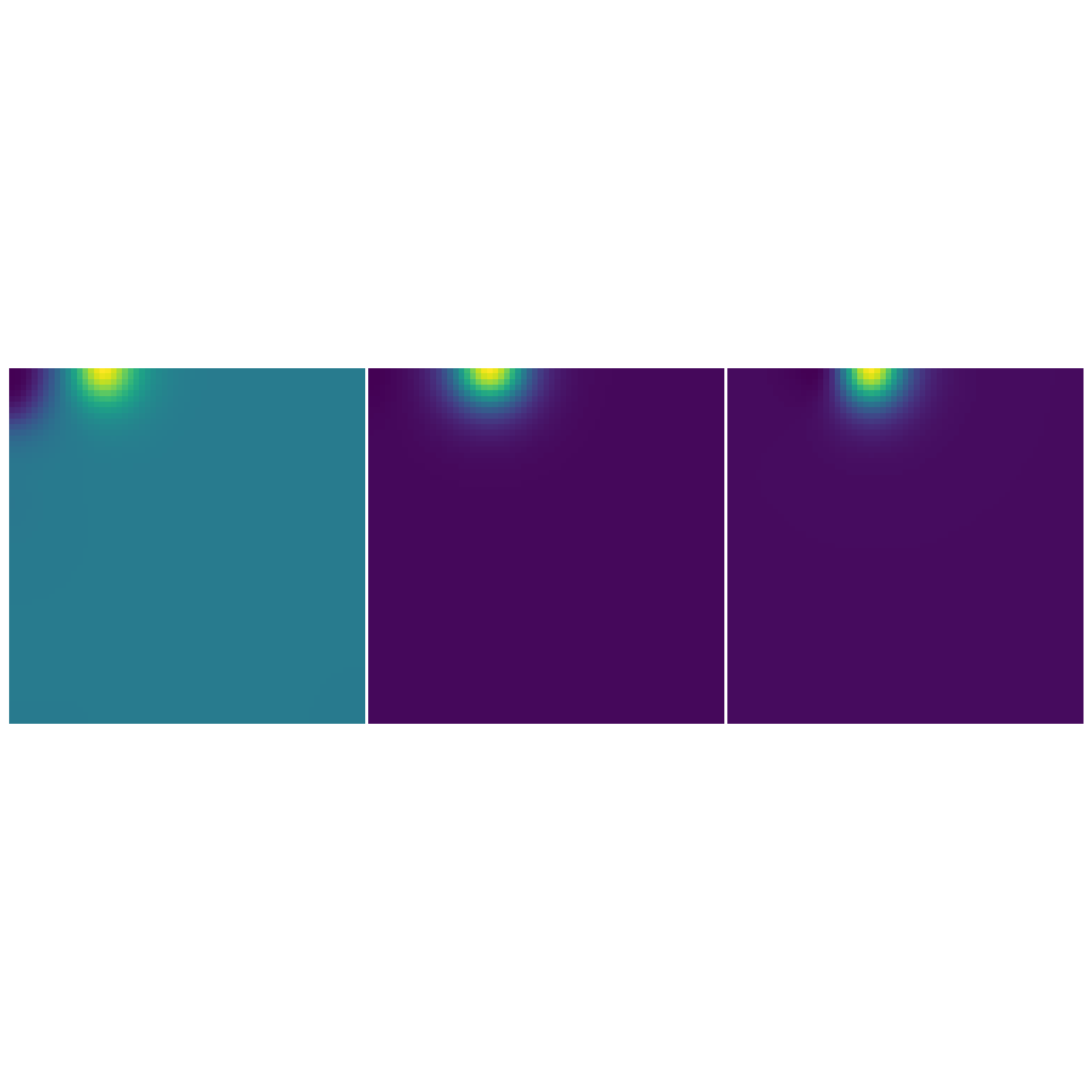}  
}
\subfigure{
    \includegraphics[width=0.3\textwidth]{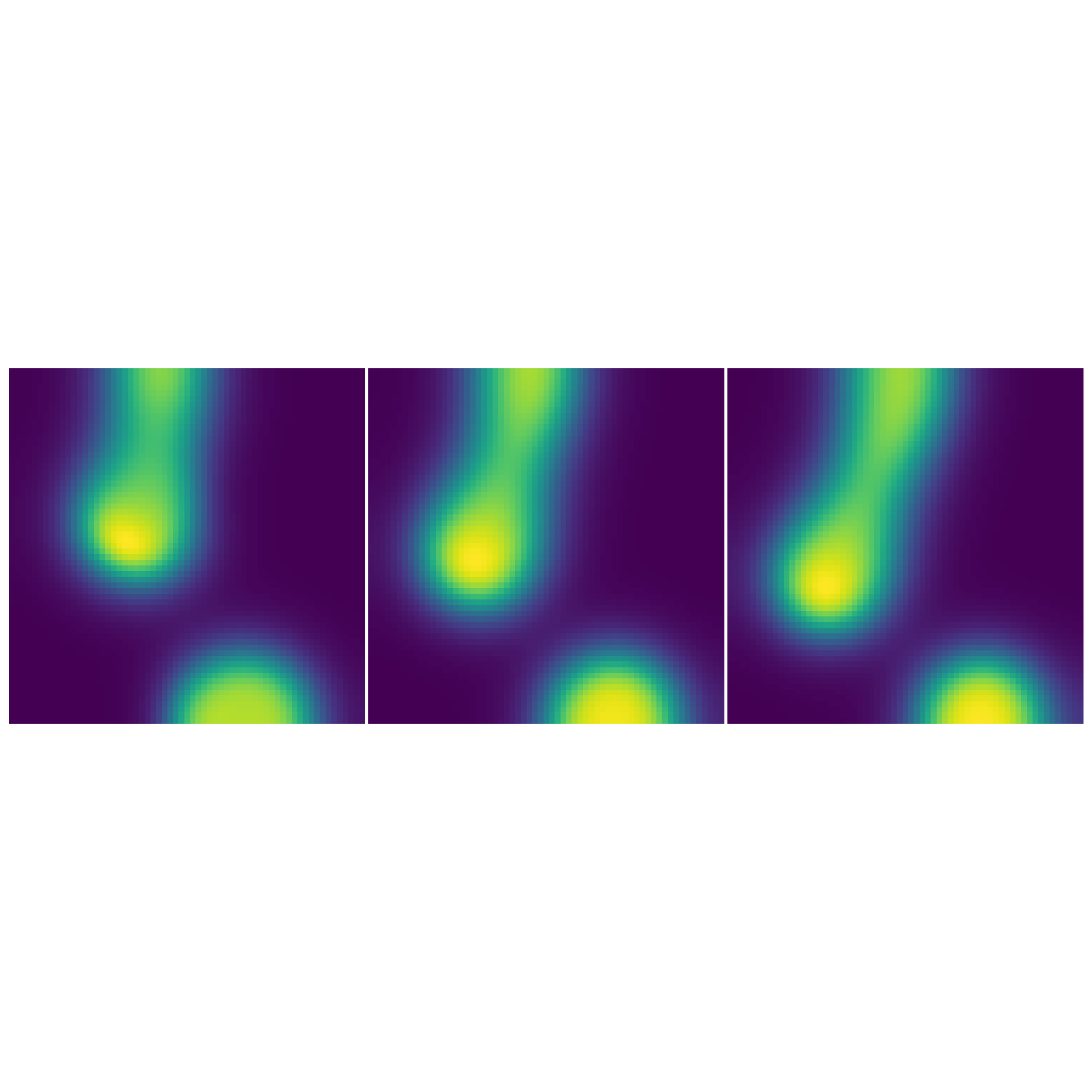}
}
\subfigure{
    \includegraphics[width=0.3\textwidth]{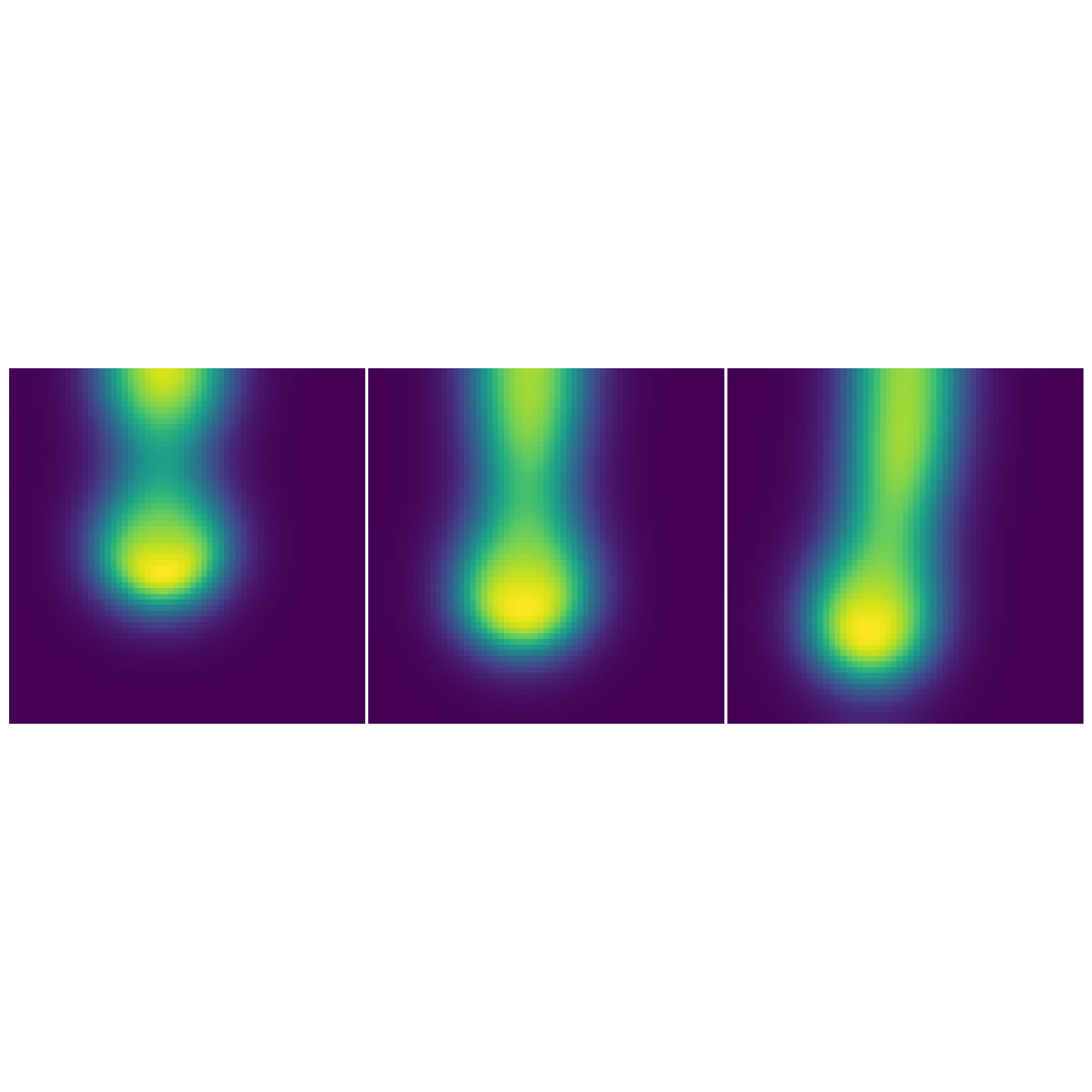}
}
\caption{Three examples depicting the evolution from the initial state to the steady state via Newton's method.}
\label{fig:evolution}
\end{figure}

\subsection{Implementations of loss functions}

\paragraph{Discrete Newton's Loss}
In solving partial differential equations (PDEs) numerically on a regular grid, the Laplace operator and other differential terms can be efficiently computed using convolution. Here, we detail the method for calculating \( J(u) \delta u - F(u) \) where \( J(u) \) is the Jacobian matrix, \( \delta u \) is the Newton step, and \( F(u) \) is the function defining the PDE.

\paragraph{Discretization}
Consider a discrete representation of a function \( u \) on a \( N \times N \) grid. The function \( u \) and its perturbation \( \delta u \) are represented as matrices:
\[
u, \delta u \in \mathbb{R}^{N \times N}
\]
The function \( F(u) \), which involves both linear and nonlinear terms, is similarly represented as \( F(u) \in \mathbb{R}^{N \times N}\).

\paragraph{Laplace Operator}
Regarding the representing the $J(u)$ with \( N \times N \) grid function $u$, the discretized Laplace operator using a finite difference method can be expressed as a convolution:
\[
-\Delta u = \begin{bmatrix} 
0 & -1 & 0 \\
-1 & 4 & -1 \\
0 & -1 & 0 
\end{bmatrix} \ast u
\]
This convolution computes the result of the Laplace operator applied to the grid function \( u \). The boundary conditions can be further incorporated in the convolution with different padding mode. Dirichlet boundary condition corresponds to zeros padding while Neumann boundary condition corresponds to replicate padding.

\subsection{Architecture of DeepONet}

A variant of \texttt{DeepONet} is used in our Newton informed neural operator. In the DeepONet, we introduce a hybrid architecture that combines convolutional layers with a static trunk basis, optimized for grid-based data inputs common in computational applications like computational biology and materials science.

\paragraph{Branch Network}
The branch network is designed to effectively downsample and process the spatial features through a series of convolutional layers:
\begin{itemize}
    \item A \textbf{Conv2D layer} with 128 filters (7x7, stride 2) initiates the feature extraction, reducing the input dimensionality while capturing coarse spatial features.
    \item This is followed by additional \textbf{Conv2D layers} (128 filters, 5x5 kernel, stride 2 and subsequently 3x3 with padding, 1x1) which further refine and compact the feature representation.
    \item The convolutional output is flattened and processed through two \textbf{fully connected layers} (256 units then down to branch features), using GELU activation.
\end{itemize}

\paragraph{Trunk Network}
The trunk utilizes a static basis represented by the tensor \texttt{V}, incorporated as a non-trainable component: The tensor \texttt{V} is precomputed, using Proper Orthogonal Decomposition (POD) as in \cite{lu2022comprehensive}, and is dimensionally compatible with the output of the branch network.

\paragraph{Forward Pass}
During the forward computation, the branch network outputs are projected onto the trunk's static basis via matrix multiplication, resulting in a feature matrix that is reshaped into the grid dimensionality for output.
\subsection*{Hyperparameters}
The following table \ref{tab:hyperparameters} summarizes the key hyperparameters used in the \texttt{DeepONet} architecture:

\begin{table}[h]
    \centering
    \begin{tabular}{|c|c|}
        \hline
        \textbf{Parameter} & \textbf{Value} \\
        \hline
        Number of Conv2D layers & 5 \\
        \hline
        Filters in Conv2D layers & 128, 128, 128, 128, 256 \\
        \hline
        Kernel sizes in Conv2D layers & 7x7, 5x5, 3x3, 1x1, 5x5 \\
        \hline
        Strides in Conv2D layers & 2, 2, 1, 1, 2 \\
        \hline
        Fully Connected Layer Sizes & 256, branch features \\
        \hline
        Activation Function & GELU \\
        \hline
    \end{tabular}
    \caption{Hyperparameters of the DeepONet  architecture}
    \label{tab:hyperparameters}
\end{table}

\subsection{Training settings}

Below we summarize the key configurations and parameters employed in the training for three cases:

\paragraph{Dataset}
\begin{itemize}
    \item \textbf{Case 1}: For method 1, we use 500 supervised data samples (with ground truth) while for method 2, we use 5000 unsupervised data samples (only with the initial state) and 500 supervised data samples.
    \item \textbf{Case 2}: For method 1, we use 5000 supervised data  while for method 2, we use 5000 unsupervised data samples and 5000 supervised data samples.
    \item  \textbf{Case 3 (Gray-Scott model)}: We only perform the method 2, with 10000 supervised data samples and 50000 unsupervised data samples.
\end{itemize}

% \subsection*{Model Architecture}
% \begin{itemize}
%     \item \textbf{Network Type}: DeepONet with 4 layers, employing features such as GELU activation and linear last layer for enhanced non-linearity handling.
%     \item \textbf{Input and Output}: Configured with a single input channel and a single output dimension, optimized for scalar field predictions.
% \end{itemize}

\paragraph{Optimization Technique}
\begin{itemize}
    \item \textbf{Optimizer}: Adam, with a learning rate of $1 \times 10^{-4}$ and a weight decay of $1 \times 10^{-6}$.
    \item \textbf{Training Epochs}: The model was trained over 1000 epochs to ensure convergence and we use \textbf{Batch Size}: 50.
\end{itemize}

These settings underscore our commitment to precision and detailed examination of neural operator efficiency in computational tasks. Our architecture and optimization choices are particularly tailored to explore and exploit the capabilities of neural networks in processing complex systems simulations.

\subsection{Benchmarking Newton's Method and neural operator based method}

\paragraph{Experimental Setup}
\label{sec:bench}
The benchmark study was conducted to evaluate the performance of a GPU-accelerated implementation of Newton's method, designed to solve systems of linear equations derived from discretizing partial differential equations. The implementation utilized CuPy with CUDA to leverage the parallel processing capabilities of the NVIDIA A100 GPU.
The hardware comprises Intel Cascade Lake 2.5 GHz CPU, an NVIDIA A100 GPU. 
% \subsection*{Hardware and Software Configuration}
% The experiments were run on a high-performance computing system equipped with the following specifications:
% \begin{itemize}
%     \item \textbf{GPU:} NVIDIA A100, leveraging CUDA for parallel computations.
%     \item \textbf{CPU:} Intel Skylake.
  
% \end{itemize}

\paragraph{Software Environment:}Ubuntu 20.04 LTS.  \textbf{Python Version:} 3.8.
    \textbf{CUDA Version:} 11.4.

% \subsection*{The implementation of Newton's method on GPU}
The Newton's method was implemented to solve the Laplacian equation over a discretized domain. Multiple system solutions were computed in parallel using CUDA streams. Key parameters of the experiment are as follows:
 \textbf{Data Type (dtype):} Single precision floating point (float32). \textbf{Number of Streams:} 10 CUDA streams to process data in parallel. \textbf{Number of Repeated Calculations:} The Newton method was executed multiples times for 500/5000 Newton linear systems, respectively, distributed evenly across the streams. \textbf{Function to Solve Systems:} The CuPy's \texttt{spsolve} was used for solving the sparse matrix systems. The following algorithm \ref{alg:newton} summary the procedure to benchmark the time used for solving Newton's system for 5000 different initial state.

\begin{algorithm}[H]
\label{alg:newton}
\caption{Solve Newton's Systems on GPU}
\begin{algorithmic}[1]
\Procedure{SolveSystemsGPU}{$A$, $u$, $rhs\_f$, $N$}
   
    \Comment{Precompute RHS and diagonal for all systems}
    \State $rhs\_list \gets rhs\_f + u^2 - A \times u$
    \State $diag\_list \gets -2 \times u$
    
    % \Comment{Transpose for efficient memory access}
    % \State $rhs\_list \gets \text{transpose}(rhs\_list)$
    % \State $diag\_list \gets \text{transpose}(diag\_list)$
    
    \Comment{Initialize solution storage}
    \State $delta\_u \gets \text{initialize zero matrix with shape of } u^T$
    
    \Comment{Solve each system}
    \For{$i = 0$ \textbf{to} $num\_sys - 1$}
        \State $rhs \gets rhs\_list[i]$
        \State $diag \gets diag\_list[i]$
        \State $J \gets A + \text{diagonal matrix with } diag \text{ on the main diagonal}$
        
        \Comment{Solve the linear system}
        \State $delta\_u[i] \gets \text{spsolve}(J, rhs)$
    \EndFor
    
    \State \textbf{return} $\text{transpose}(delta\_u)$
\EndProcedure
\end{algorithmic}
\end{algorithm}

\section{Supplemental material for proof}\subsection{Preliminaries}\label{pre}
\begin{defi}[Sobolev Spaces \cite{evans2022partial}]
			Let $\Omega$ be $[0,1]^d$ and let $D$ be the operator of the weak derivative of a single variable function and $D^{\boldsymbol{\alpha}}=D^{\alpha_1}_1D^{\alpha_2}_2\ldots D^{\alpha_d}_d$ be the partial derivative where $\boldsymbol{\alpha}=[\alpha_{1},\alpha_{2},\ldots,\alpha_d]^T$ and $D_i$ is the derivative in the $i$-th variable. Let $n\in\sN$ and $1\le p\le \infty$. Then we define Sobolev spaces\[W^{n, p}(\Omega):=\left\{f \in L^p(\Omega): D^{\boldsymbol{\alpha}} f \in L^p(\Omega) \text { for all } \boldsymbol{\alpha} \in \sN^d \text { with }|\boldsymbol{\alpha}| \leq n\right\}\] with a norm \[\|f\|_{W^{n, p}(\Omega)}:=\left(\sum_{0 \leq|\alpha| \leq n}\left\|D^{\alpha} f\right\|_{L^p(\Omega)}^p\right)^{1 / p}\] if $p<\infty$, and $\|f\|_{W^{n, \infty}(\Omega)}:=\max_{0 \leq|\alpha| \leq n}\left\|D^{\alpha} f\right\|_{L^\infty(\Omega)}$.

   Furthermore, for $\boldsymbol{f}=(f_1,\ldots,f_d)$, $\boldsymbol{f}\in W^{1,\infty}(\Omega,\sR^d)$ if and only if $ f_i\in W^{1,\infty}(\Omega)$ for each $i=1,2,\ldots,d$ and \[\|\boldsymbol{f}\|_{W^{1,\infty}(\Omega,\sR^d)}:=\max_{i=1,\ldots,d}\{\|f_i\|_{W^{1,\infty}(\Omega)}\}.\] When $p=2$, denote $W^{n,2}(\Omega)$ as $H^n(\Omega)$ for $n\in\sN_+$.
		\end{defi}

  \begin{prop}[\cite{mhaskar1996neural}]\label{opeartor}
Suppose $\sigma$ is a is a continuous non-polynomial function and $K$ is a compact in $\sR^d$, then there are  positive integers $p$,  constants $w_k, \zeta_{k}$ for $k=1, \ldots, p$ and bounded linear functionals $c_k:H^{r}(K)\to \sR$ such that for any $v\in H^{r}(K)$,
\begin{equation}
	\left\|v-\sum_{k=1}^{p} c_{k}(v) \sigma\left(\vw_{k}\cdot \vx+\zeta_k\right)\right\|_{L^{2}(K)} \leq c p^{-r / d}\|v\|_{H^{r}(K)}.
\end{equation}
\end{prop}

  \begin{prop}[\cite{poggio2017and,yang2023nearly}]\label{function}
Suppose $\sigma$ is a continuous non-polynomial function and $\Omega$ is a compact subset of $\sR^d$. For any Lipschitz-continuous function $f$, there exists a shallow neural network such that
\begin{equation}
	\left\|f-\sum_{j=1}^{m} a_{j} \sigma\left(\vomega_{j}\cdot \vx+b_j\right)\right\|_{\infty} \leq C m^{-1 / d},
\end{equation}
where $C$ depends on the Lipschitz constant but is independent of $m$.
\end{prop}

\begin{lem}[\cite{lanthaler2022error}]\label{cover}
The $\epsilon$-covering number of $[-B, B]^d$, $K(\epsilon)$, satisfies
\[
K(\epsilon) \leqslant \left(\frac{C B}{\epsilon}\right)^d,
\]
for some constant $C > 0$, independent of $\epsilon$, $B$, and $d$.
\end{lem}

\textbf{Step 5:} Now we estimate 

\subsection{Proof of Theorem \ref{approximation}}
\begin{figure}[h!]
\centering
\includegraphics[scale=0.52]{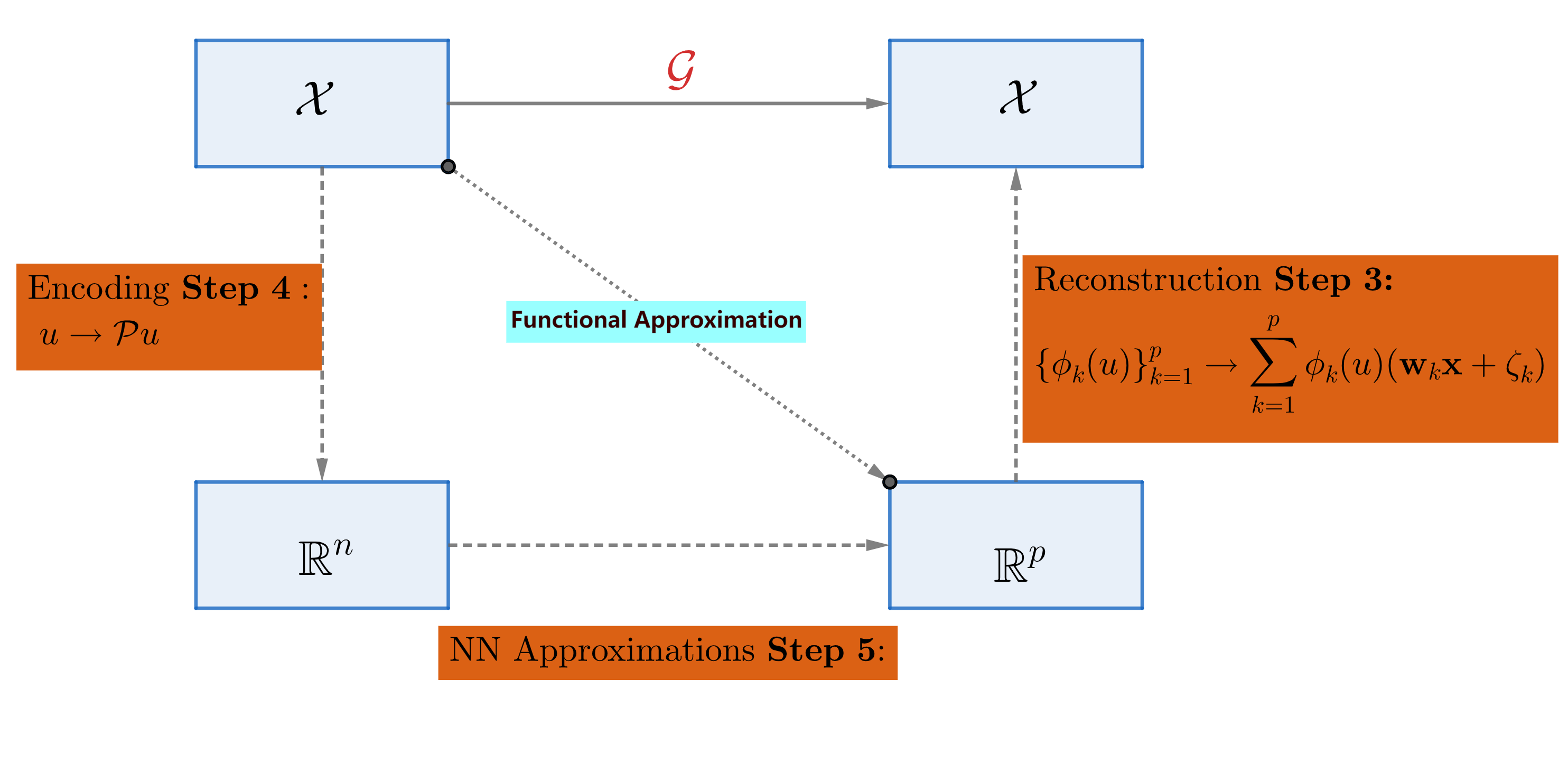}
\caption{The sketch of proof for Theorem \ref{approximation}. The details of each step are provided in the following.}
\label{sketch app}
\end{figure}

In this subsection, we present the proof of Theorem \ref{approximation}, which describes the approximation ability of DeepONet. The sketch of the proof is illustrated in Fig.~\ref{sketch app}.

\begin{proof}[Proof of Theorem \ref{approximation}]
    \textbf{Step 1:} Firstly, we need to verify that the target operator $\fG(u)$ is well-defined.

Due to Assumption \ref{app assump} (i) and \cite[Theorem 6 in Section 6.2]{evans2022partial}, we know that for $u\in\fX\subset H^2(\Omega)$, Equation (\ref{linear}) will have unique solutions. This means that $\fG(u)$ is a well-defined operator for the input space $u\in\fX$.

\textbf{Step 2:} Secondly, we aim to verify that $\fG(u)$ is a Lipschitz-continuous operator in $H^2(\Omega)$ for $u\in\fX$.

Consider the following:
\begin{align}
f'(u+v) &= f'(u) + vf''(\xi_1)\notag \\
f(u+v) &= f(u) + vf'(\xi_2)\notag \\
\delta u_v(\vx) &= \delta u(\vx) + \epsilon(\vx)
\end{align}
where $\delta u(\vx)$ is the solution of Eq.(\ref{linear}) for the input $u$, and $\delta_v u(\vx)$ is the solution of Eq.(\ref{linear}) for the input $u+v$. 
Denote \[\delta_v u(\vx)-\delta u(\vx)=:\epsilon(\vx).\]

Therefore, we have:
\begin{equation}
\begin{cases}
(\fL-f'(u+v))\epsilon(\vx) = \Delta v - v(f'(\xi_2) + f''(\xi_1)\delta u), & \vx \in \Omega \\
\epsilon(\vx) = 0, & \vx \in \partial\Omega.
\end{cases}
\label{per}
\end{equation}

Since $u$ and $v$ are in $H^2$ and $\partial\Omega$ is in $C^2$ (Assumption \ref{app assump} (iii)), according to \cite[Theorem 4 in Section 6.3]{evans2022partial}, there exist constants $C$ $\footnote{In this paper, we consistently employ the symbol $C$ as a constant, which may vary from line to line.}$ and $\bar{C}$ such that:
\begin{align}
\|\epsilon(\vx)\|_{H^2(\Omega)} &\le C\|\fL v - v(f'(\xi_2) + f''(\xi_1)\delta u)\|_{L^2(\Omega)}\notag \\
&\le \bar{C} \|v(\vx)\|_{H^2(\Omega)}.
\end{align}
The last inequality is due to the boundedness of $f'(\xi_2) + f''(\xi_1)\delta u$ (Assumption \ref{app assump} (ii)).

\textbf{Step 3:} In the approximation, we first reduce the operator learning to functional learning. 

When the input function $u(\vx)$ belongs to $\fX \subset H^2$, the output function $\delta u$ also belongs to $H^2$, provided that $\partial \Omega$ is of class $C^2$. The function $\fG(u) = \delta u$ can be approximated by a two-layer network architected by the activation function $\sigma(x)$, which is not a polynomial, in the following form by Proposition \ref{opeartor} \cite{mhaskar1996neural} (given in Subsection \ref{per}):
\begin{equation}
    \left\| \fG(u)(\vx) - \sum_{k=1}^{p} c_{k}[\fG(u)] \sigma\left( \vw_{k} \cdot \vx + \zeta_k \right) \right\|_{L^2(\Omega)} \leq C_1 p^{-\frac{2}{d}} \| \fG(u) \|_{H^2(\Omega)}, \label{step1}
\end{equation}
where $\vw_k \in \mathbb{R}^{d}$, $\zeta_{k} \in \mathbb{R}$ for $k = 1, \ldots, p$, $c_k$ is a continuous functional, and $C_1$ is a constant independent of the parameters.

Denote $\phi_k(u) = c_{k}[\fG(u)]$, which is a Lipschitz-continuous functional from $H^2(\Omega)$ to $\mathbb{R}$, which is due to $\fG$ is a Lipschitz-continuous operator and $c_k$ is a linear functional. The remaining task in approximation is to approximate these functionals by neural networks.

\textbf{Step 4:} In this step, we reduce the functional learning to function learning by applying the operator $\fP$ as in Assumption \ref{app assump} (iv).

Based on $\phi_k(u)$ being a Lipschitz-continuous functional in $H^2(\Omega)$, we have
\[ |\phi_k(u) - \phi_k(\fP u)| \le L_k \| u - \fP u \|_{H^2(\Omega)} \le L_k \epsilon, \]
where $L_k$ is the Lipschitz constant of $\phi_k(u)$ for $u \in \fX$.

Furthermore, since $\fP u$ is an $n$-dimensional term, i.e., it can be denoted by the $n$-dimensional vector $\bar{\fP} u \in \mathbb{R}^n$, we can rewrite $\phi_k(\fP u)$ as $\psi_k(\bar{\fP} u)$, where $\psi_k: \mathbb{R}^n \to \mathbb{R}$ for $l=1,\ldots,p$. Furthermore, $\psi_k$ is a Lipschitz-continuous function since $\phi_k$ is Lipschitz-continuous and $\fP$ is a continuous linear operator.

\textbf{Step 5:} In this step, we will approximate $\psi_k$ using shallow neural networks.

Due to Proposition \ref{function}, we have that there is a shallow neural network such that
\begin{equation}
	\left\|\psi_k(\bar{\fP}u)- \vA_{k} \sigma\left(\vM_k\cdot \bar{\fP}u+\vb_k\right)\right\|_{\infty} \leq C m^{-1 / d},
\end{equation}
where $\va_k^\T\in\sR^{ m}$, $\vM_k\in\sR^{m\times n}$, and $\vb_k\in\sR^{ m}$. For the simplicity notations, we can replace $\vM_k\cdot \bar{\fP}$ by a operator $\fW_k$.

Above all, we have that there is a neural network in $\Xi_p$ such that
\begin{equation}
\left\|\sum_{k=1}^{p} \vA_{k} \sigma\left(\fW_k u+\vb_k\right) \sigma\left( \vw_{k} \cdot \vx + \zeta_k \right)-\fG(u)\right\|_{L^2(\Omega)}\le C_1m^{-\frac{1}{ n}}+C_2(\epsilon+p^{-\frac{2}{d}}),
\end{equation}
where $C_1$ is a constant independent of $m$, $\epsilon$, and $p$, $C_2$ is a constant depended on $p$.

\end{proof}

\begin{figure}[h]
    \centering 
    \includegraphics[width=0.57\textwidth]{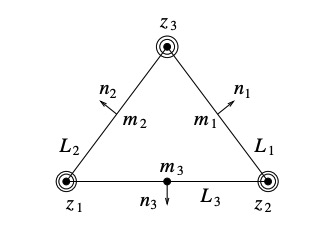} 
    \caption{Argyris method} 
    \label{arg}
\end{figure}

Here, we discuss more about the embedding operator $\fP$. One approach is to use the Argyris element \cite{brenner2008mathematical}. This method involves considering the $21$ degrees of freedom shown in Fig.~\ref{arg}. In this figure, each $\bullet$ denotes evaluation at a point, the inner circle represents evaluation of the gradient at the center, and the outer circle denotes evaluation of the three second derivatives at the center. The arrows indicate the evaluation of the normal derivatives at the three midpoints.

Another alternative approach to discretizing the input space is to use the bi-cubic Hermite finite element method \cite{li2002new,hu2015minimal}.

\subsection{Proof of Theorem \ref{genL2}}\label{gen}
The proof of Theorem \ref{genL2} is inspired by that in \cite{lanthaler2022error}.
 \begin{proof}[Proof of Theorem \ref{genL2}]
     \textbf{Step 1:} To begin with, we introduce a new term called the middle term, denoted as $\fE_{Sm}(\vtheta)$, defined as follows:
\[
\fE_{Sm}(\vtheta) := \frac{1}{M_u} \sum_{j=1}^{M_u} \int_{\Omega} \left| \fG(u_j)(\vx) - \fO(u_j;\vtheta)(\vx) \right|^2 \D \vx,
\]
This term represents the limit case of $\fE_{S}(\vtheta)$ as the number of samples in the domain of the output space tends to infinity ($M_x \to \infty$).

Then the error can be divided into two parts:\begin{equation}
    |\rmE(\fE_{S}(\vtheta)-\fE_{Sc}(\vtheta)|\le |\rmE(\fE_{S}(\vtheta)-\fE_{Sm}(\vtheta)|+|\rmE(\fE_{Sm}(\vtheta)-\fE_{Sc}(\vtheta)|.
\end{equation}

\textbf{Step 2:} For $|\rmE(\fE_{Sm}(\vtheta)-\fE_{Sc}(\vtheta)|$, this is the classical generalization error analysis, and the result can be obtained from \cite{schmidt2020nonparametric,yang2024deeper}. We omit the details of this part, which can be expressed as
\begin{equation}
    |\rmE(\fE_{Sm}(\vtheta)-\fE_{Sc}(\vtheta))| \le \frac{C d_{\vtheta}\log M_x}{M_x},
\end{equation}
where $C$ is independent of the number of parameters $d_{\vtheta}$ and the sample size $M_x$. In the following steps, we are going to estimate $|\rmE(\fE_{S}(\vtheta)-\fE_{Sm}(\vtheta))|$, which is the error that comes from the sampling of the input space of the operator.

\textbf{Step 3: } Denote 
\[
S_{\vtheta}^M:=\frac{1}{M} \sum_{j=1}^{M} \int_{\Omega} \left| \fG(u_j)(\vx) - \fO(u_j;\vtheta)(\vx) \right|^2 \D \vx.
\] 
We first estimate the gap between $S_{\vtheta}^M$ and $S_{\vtheta'}^M$ for any bounded parameters $\vtheta,\vtheta'$. Due to Assumption \ref{gen app} (i) and (ii), we have that 
\begin{align}
    &|S_{\vtheta}^M-S_{\vtheta'}^M|\notag\\
    \le& \frac{1}{M}\sum_{j=1}^M\left|\int_{\Omega} \left| \fG(u_j)(\vx) - \fO(u_j;\vtheta)(\vx) \right|^2-\left| \fG(u_j)(\vx) - \fO(u_j;\vtheta')(\vx) \right|^2 \D \vx\right|\notag\\
    \le& \frac{1}{M}\sum_{j=1}^M\left|\int_{\Omega} \left| 2\fG(u_j)(\vx) +\fO(u_j;\vtheta)(\vx)+\fO(u_j;\vtheta')(\vx)\right|\cdot\left| \fO(u_j;\vtheta)(\vx) - \fO(u_j;\vtheta')(\vx) \right| \D \vx\right|\notag\\
    \le &   \frac{4}{M}\sum_{j=1}^M\Psi(u_j) \Phi(u_j)\cdot\left\|\vtheta-\vtheta^{\prime}\right\|_{\ell^{\infty}}.
\end{align}

\textbf{Step 4:} Based on Step 3, we are going to estimate 
\[
\rmE \left[\sup_{\vtheta\in[-B,B]^{d_{\vtheta}}}\left|S_{\vtheta}^M-\rmE S_{\vtheta}^M\right|^p\right]^{\frac{1}{p}}
\] 
by covering the number of the spaces.

Set $\{\vtheta_1, \ldots, \vtheta_K\}$ is a $\varepsilon$-covering of $\in[-B, B]^{d_{\vtheta}}$ i.e. for any $\vtheta$ in $\in[-B, B]^{d_{\vtheta}}$, there exists $j$ with $\left\|\vtheta-\vtheta_j\right\|_{\ell _\infty} \leqslant \epsilon$. Then we have \begin{align}
& \rmE\left[\sup _{\vtheta \in[-B, B]^d}\left|S_{\vtheta}^{M}-\rmE\left[S_{\vtheta}^{M}\right]\right|^p\right]^{1 / p} \notag\\\le& \rmE {\left[\left(\sup _{\vtheta \in[-B, B]^d}\left|S_{\vtheta}^{M}-S_{\vtheta_{j(\vtheta)}}^{M}\right|+\left|S_{j(\vtheta)}^{M}-\rmE\left[S_{j(\vtheta)}^{M}\right]\right|\right.\right.} \left.\left.+\left|\rmE\left[S_{\vtheta_{j(\vtheta)}}^{M}\right]-\rmE\left[S_{\vtheta}^{M}\right]\right|\right)^p\right]^{1 / p}  \notag\\\le& \rmE\left[\left(\max _{j=1, \ldots, K}\left|S_{j(\vtheta)}^{M}-\rmE\left[S_{j(\vtheta)}^{M}\right]\right|\right.\right. \left.\left.+\frac{8 \epsilon}{M}\left(\sum_{j=1}^{M}\left|\Psi\left(u\right)\right|\left|\Phi\left(u\right)\right|\right)\right)^p\right]^{1 / p}\notag\\\le&  8 \epsilon \rmE\left[|\Psi \Phi|^p\right]^{1 / p}+\rmE\left[\max _{j=1, \ldots, K}\left|S_{\vtheta_j}^{M}-\rmE\left[S_{\vtheta_j}^{M}\right]\right|^p\right]^{1 / p} .
\end{align}

For $8 \epsilon \rmE\left[|\Psi \Phi|^p\right]^{1 / p}$, it can be approximate by \[8 \epsilon \rmE\left[|\Psi \Phi|^p\right]^{1 / p} \leqslant 8 \epsilon \rmE\left[|\Psi|^{2 p}\right]^{1 / 2 p} \rmE\left[|\Phi|^{2 p}\right]^{1 / 2 p}=8 \epsilon\|\Psi\|_{L^{2 p}}\|\Phi\|_{L^{2 p}} .\]

For $\rmE\left[\max _{j=1, \ldots, K}\left|S_{\vtheta_j}^{M}-\rmE\left[S_{\vtheta_j}^{M}\right]\right|^p\right]^{1 / p} $, by applied the result in \cite{welti2020high,lanthaler2022error}, we know \[\rmE\left[\max _{j=1, \ldots, K}\left|S_{{\vtheta}_j}^{M}-\rmE\left[S_{{\vtheta}_j}^{M}\right]\right|^p\right]^{1 / p} \le \frac{16 K^{1 / p} \sqrt{p}\|\Psi\|_{L^{2 p}}^2}{\sqrt{M}} .\]

\textbf{Step 5:} Now we estimate $|\rmE(\fE_{S}(\vtheta)-\fE_{Sm}(\vtheta)|$.

Due to Assumption \ref{gen app} and directly calculation, we have that $$
\|\Psi\|_{L^{2 p}},\|\Phi\|_{L^{2 p}} \leqslant C(1+\gamma \kappa p)^\kappa,
$$
for constants $C, \gamma>0$, depending only the measure $\mu$ and the constant $C$ appearing in the upper bound (\ref{kappa}).

Based on Lemma \ref{cover}, we have that \[
\rmE\left[\sup _{{\vtheta} \in[-B, B]^{d_{\vtheta}}}\left|S_{\vtheta}^{M_u}-\rmE\left[S_{\vtheta}^{M_u}\right]\right|^p\right]^{1 / p} \leqslant 16 C^2(1+\gamma \kappa p)^{2 \kappa}\left(\epsilon+\left(\frac{C B}{\epsilon}\right)^{d_{\vtheta} / p} \frac{\sqrt{p}}{\sqrt{M_u}}\right),
\]
for some constants $C, \gamma>0$, independent of $\kappa, \mu, B, d_{\vtheta}, N, \epsilon>0$ and $p \geqslant 2$. We now choose $\epsilon=\frac{1}{\sqrt{M_u}}$, so that
$$
\epsilon+\left(\frac{C B}{\epsilon}\right)^{d_{\vtheta} / p} \frac{\sqrt{p}}{\sqrt{M_u}}=\frac{1}{\sqrt{M_u}}\left(1+(C B \sqrt{M_u})^{d_{\vtheta} / p} \sqrt{p}\right) .
$$

Next, let $p=d_{\vtheta} \log (C B \sqrt{M_u})$. Then,
$$
(C B \sqrt{M_u})^{d_{\vtheta} / p} \sqrt{p}=\exp \left(\frac{\log (C B \sqrt{M_u}) d_{\vtheta}}{p}\right) \sqrt{p}=e \sqrt{d_{\vtheta} \log (C B \sqrt{M_u})},
$$
and thus we conclude that
$$
\epsilon+\left(\frac{C B}{\epsilon}\right)^{d_{\vtheta} / p} \frac{\sqrt{p}}{\sqrt{M_u}} \leqslant \frac{1}{\sqrt{M_u}}\left(1+e \sqrt{d_{\vtheta} \log (C B \sqrt{M_u})} .\right) .
$$

On the other hand, we have
$$
(1+\gamma \kappa p)^{2 \kappa}=\left(1+\gamma \kappa d_{\vtheta} \log (C B \sqrt{M_u})\right)^{2 \kappa} .
$$

Increasing the constant $C>0$, if necessary, we can further estimate
$$
\left(1+\gamma \kappa d_{\vtheta} \log (C B \sqrt{M_u})\right)^{2 \kappa}\left(1+e \sqrt{d_{\vtheta} \log (C B \sqrt{M_u})} .\right) \leqslant C\left(1+d_{\vtheta} \log (C B \sqrt{M_u})\right)^{2 \kappa+1 / 2},
$$
where $C>0$ depends on $\kappa, \gamma, \mu$ and the constant appearing in (\ref{kappa}), but is independent of $d_{\vtheta}, B$ and $N$. We can express this dependence in the form $C=C(\mu, \Psi, \Phi)>0$, as the constants $\kappa$ and $\gamma$ depend on the Gaussian tail of $\mu$ and the upper bound on $\Psi, \Phi$.

Therefore, \begin{align}
        |\rmE(\fE_{S}(\vtheta)-\fE_{Sm}(\vtheta)| &\le \sup _{\vtheta \in[-B, B]^{d_{\vtheta}}}|S_{\vtheta}^{M_u}-\rmE\left[S_{\vtheta}^{M_u}\right]|\notag\\&\le \frac{C}{\sqrt{M_u}}\left(1+C d_{\vtheta} \log (C B \sqrt{M_u})^{2 \kappa+1 / 2}\right).
    \end{align}

 \end{proof}

%%%%%%%%%%%%%%%%%%%%%%%%%%%%%%%%%%%%%%%%%%%%%%%%%%%%%%%%%%%%

\end{document}